\pdfoutput=1

\documentclass[11pt]{article}

\usepackage[]{acl}

\usepackage{times}
\usepackage{latexsym}

\usepackage[T1]{fontenc}

\usepackage[utf8]{inputenc}

\usepackage{microtype}

\usepackage{inconsolata}

\usepackage{graphicx}

\usepackage{subcaption}
\usepackage{graphicx}
\usepackage{amsmath}
\usepackage{verbatim}
\usepackage{algorithm}
\usepackage{algpseudocode}
\usepackage{amsmath}
\usepackage{amssymb}
\usepackage{multirow}
\usepackage{makecell}
\usepackage{booktabs}
\usepackage{stfloats}
\usepackage[misc]{ifsym}
\usepackage{CJKutf8}
\usepackage{color}
\usepackage{bbding}
\usepackage{wrapfig}
\usepackage{xspace} 
\usepackage{enumitem}
\usepackage{hyperref}
\usepackage{pifont}
\usepackage{tcolorbox}
\usepackage{tabularx}
\usepackage{longtable}
\tcbuselibrary{breakable}
\usepackage{xcolor}
\usepackage{colortbl}

\newcommand*\samethanks[1][\value{footnote}]{\footnotemark[#1]}

\newcommand{\website}{\includegraphics[height=1em,trim=0 0em 0 0]{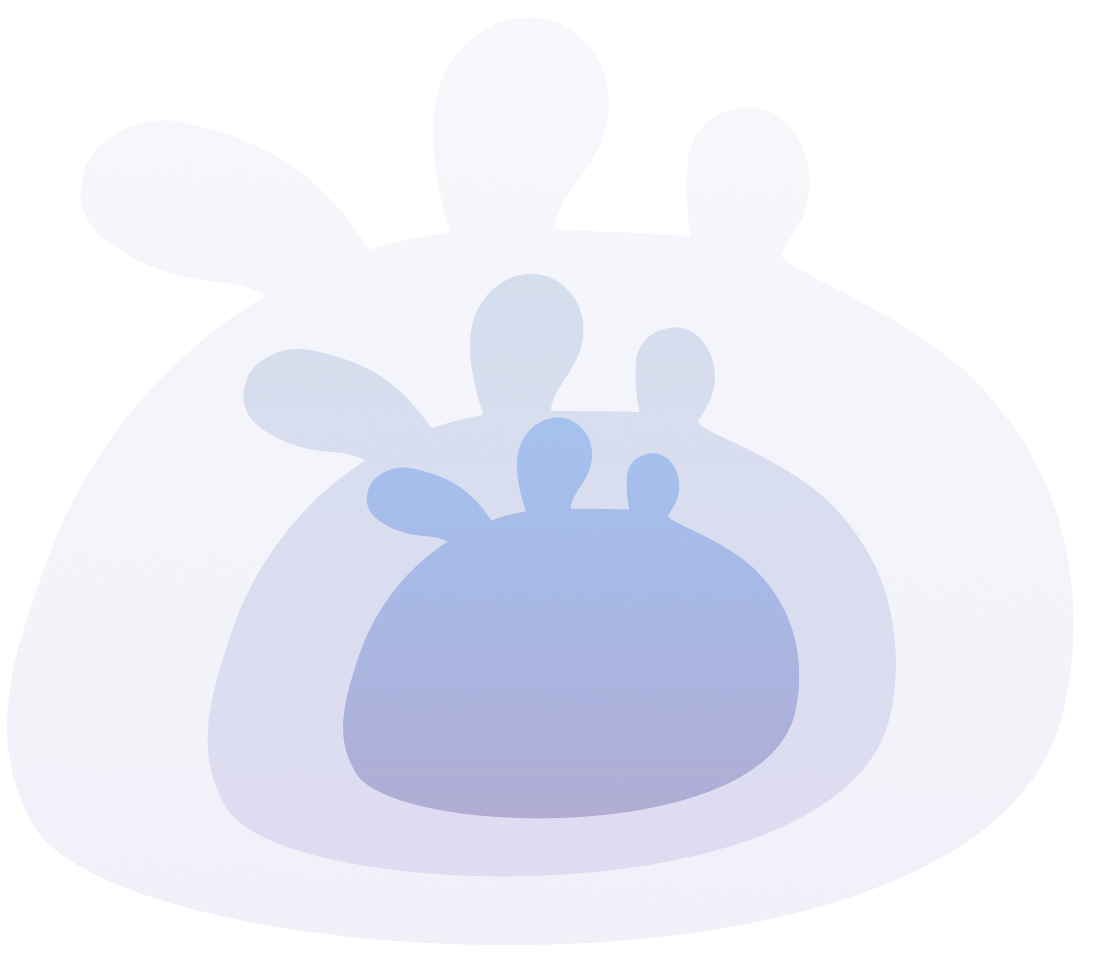}}

\newcommand{\datasetemoji}{\includegraphics[height=1em,trim=0 0em 0 0]{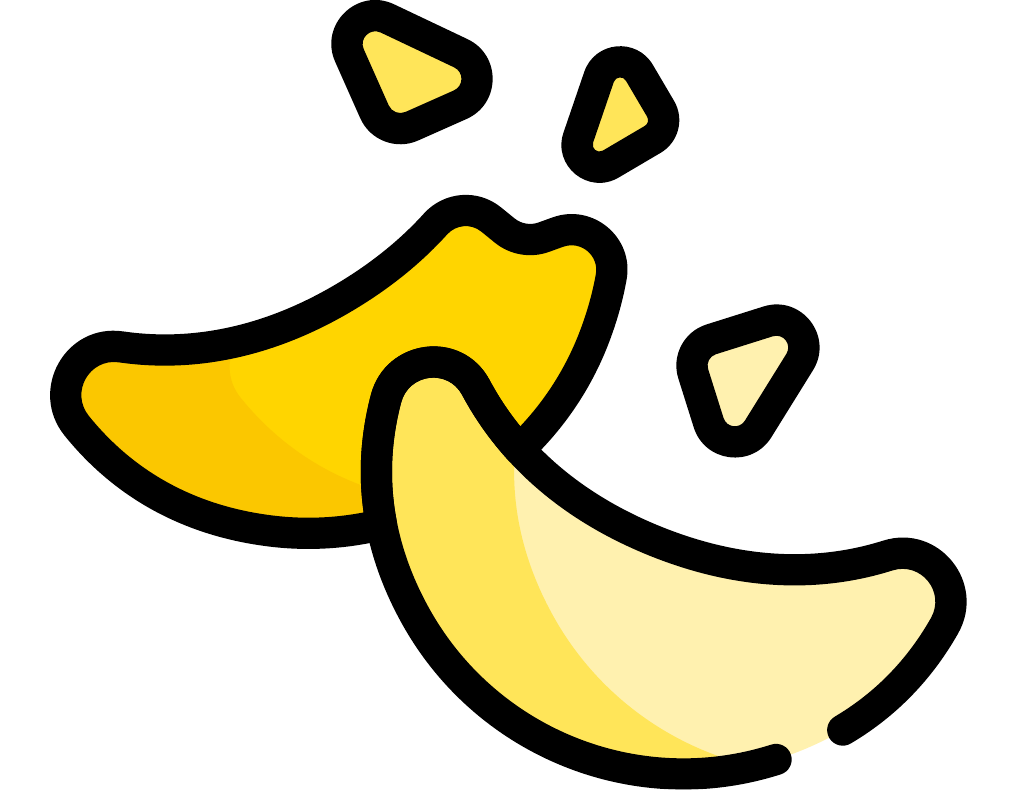}}
\newcommand{\datasetlogo}{\datasetemoji\textsc{Crisp}\xspace}

\newcommand{\llmemoji}{\includegraphics[height=1em,trim=0 0em 0 0]{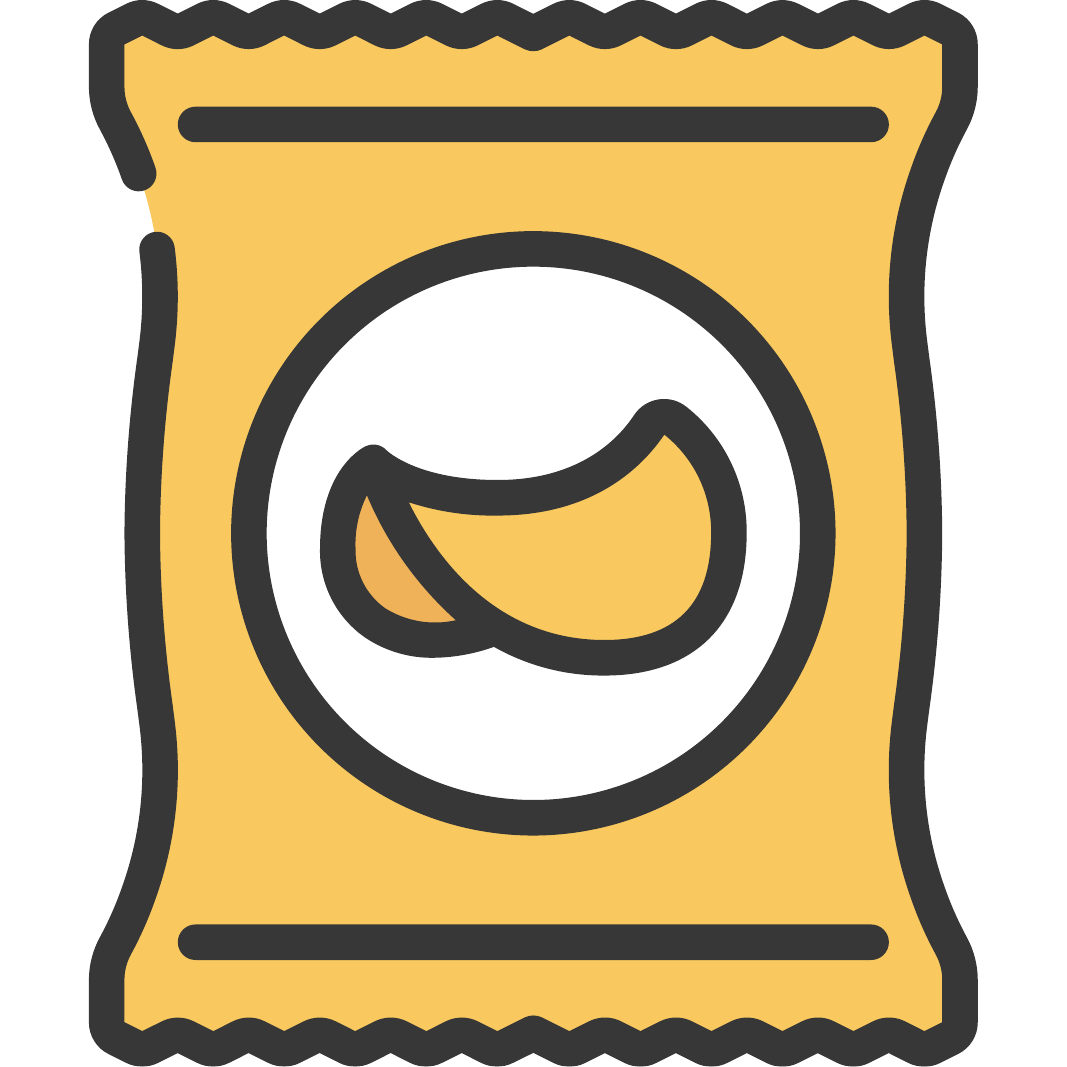}}
\newcommand{\llmlogo}{\llmemoji\textsc{Crispers}\xspace}

\definecolor{identification}{RGB}{234,231,242}
\definecolor{restructuring}{RGB}{214,225,241}

\title{\datasetlogo: Cognitive Restructuring of Negative Thoughts through \\ Multi-turn Supportive Dialogues}

\author{
    Jinfeng Zhou\textsuperscript{\rm 1}\thanks{Equal contribution.} \quad
    Yuxuan Chen\textsuperscript{\rm 1}\samethanks{} \quad
    Jianing Yin\textsuperscript{\rm 2} \quad
    Yongkang Huang\textsuperscript{\rm 3} \quad
    Yihan Shi\textsuperscript{\rm 4} \\
    \textbf{Xikun Zhang\textsuperscript{\rm 5} \quad
    Libiao Peng\textsuperscript{\rm 3} \quad
    Rongsheng Zhang\textsuperscript{\rm 6} \quad
    Tangjie Lv\textsuperscript{\rm 6}} \\
    \textbf{Zhipeng Hu\textsuperscript{\rm 6} \quad
    Hongning Wang\textsuperscript{\rm 1} \quad
    Minlie Huang\textsuperscript{\rm 1}}\thanks{Corresponding author.} \\
    \textsuperscript{\rm 1}The CoAI Group, DCST, Tsinghua University \quad 
    \textsuperscript{\rm 2}University of Pennsylvania \\
    \textsuperscript{\rm 3}Lingxin AI \quad
    \textsuperscript{\rm 4}Harvard Graduate School of Education, Harvard University \\
    \textsuperscript{\rm 5}Department of Psychology and Behavioral Sciences, Zhejiang University \quad
    \textsuperscript{\rm 6}Fuxi AI Lab \\
    \texttt{zjf23@mails.tsinghua.edu.cn \quad aihuang@tsinghua.edu.cn} \\
    \website \ \url{https://peppy-ai.com/}
}

\begin{document}

\maketitle

\begin{abstract}

Cognitive Restructuring (CR) is a psychotherapeutic process aimed at identifying and restructuring an individual’s negative thoughts, arising from mental health challenges, into more helpful and positive ones via multi-turn dialogues. Clinician shortage and stigma urge the development of human-LLM interactive psychotherapy for CR. Yet, existing efforts implement CR via simple text rewriting, fixed-pattern dialogues, or a one-shot CR workflow, failing to align with the psychotherapeutic process for effective CR. To address this gap, we propose \textsc{CRDial}, a novel framework for CR, which creates multi-turn dialogues with specifically designed identification and restructuring stages of negative thoughts, integrates sentence-level supportive conversation strategies, and adopts a multi-channel loop mechanism to enable iterative CR. With \textsc{CRDial}, we distill \datasetlogo, a large-scale and high-quality bilingual dialogue dataset, from LLM. We then train \llmlogo, \textsc{Crisp}-based conversational LLMs for CR, at 7B and 14B scales. Extensive human studies show the superiority of \textsc{Crispers} in pointwise, pairwise, and intervention evaluations.\footnote{Repository: \url{https://github.com/thu-coai/Crisp}}

\end{abstract}

\section{Introduction}
\label{sec:introduction}

\begin{figure}[t]
    \centering
    \includegraphics[width=\columnwidth]{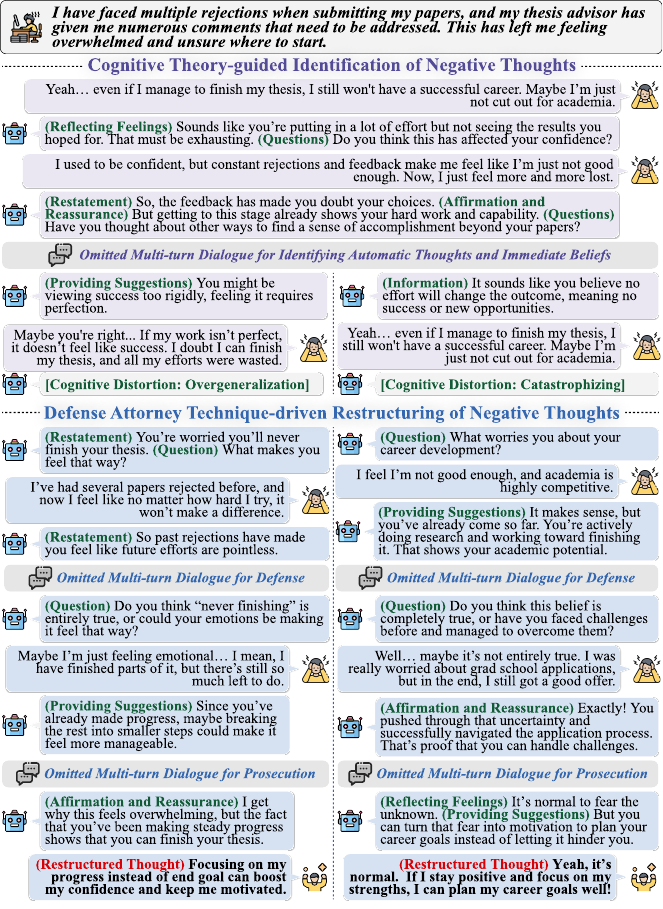}
    \caption{A data example of \textsc{Crsip} crafted via LLMs using \textsc{CRDial}, which \colorbox{identification}{\textit{identifies}} and \colorbox{restructuring}{\textit{restructures}} multiple negative thoughts (i.e., \textit{cognitive distortions}) via multi-turn dialogue with emotional support.}
    \label{intro_case}
\vspace{-4mm}
\end{figure}

Negative thoughts arise from cognitive distortions that create a distorted view of reality and are entrenched in individuals facing mental health challenges, which may lead to depression and even suicidal ideation \cite{ct}. 
A proven intervention for such thoughts is \textbf{Cognitive Restructuring} (\textbf{CR}, \citealt{cbt}). As shown in Figure \ref{intro_case}, CR is a therapeutic process where a therapist would first initiate a dialogue from an individual's negative situation (e.g., ``\textit{receiving numerous comments from his/her advisor}'') and progressively explores their thoughts from surface (e.g., ``\textit{I just feel more and more lost}'') to the core negative thought (e.g., \textbf{catastrophizing}: ``\textit{Even if I finish the thesis, I still won't have a successful career}''). The therapist should then apply restructuring techniques to transform this catastrophizing thought into a more helpful and positive one (e.g., ``\textit{I believe this fear is normal and will motivate me to plan my career goals}'').

Yet, barriers like clinician shortages and stigma often limit people's access to therapists \cite{therapy_shortage}, urging the development of human-machine interactive psychotherapy for CR.
But the process of CR often faces several key challenges \cite{ct} which make it non-trivial for machines to successfully carry out CR.
\textbf{(1) Cognitive persistence}: Cognitive distortions are rooted in individuals' thought patterns, impeding their self-awareness of negative thoughts.
\textbf{(2) Emotional management}: Engaging in CR may cause individuals to feel discomfort or resist changing their entrenched thought patterns, thus evoking intense emotions and disrupting the therapeutic process.
\textbf{(3) Individual differences}: Individuals' personalities shape their cognitive styles, resulting in diverse thought patterns. For example, in Figure \ref{intro_case}, an individual may also experience cognitive distortions like \textit{overgeneralization} (``\textit{all my efforts were wasted}'') or \textit{catastrophizing} (``\textit{I still won’t have a successful career}''), or even both.
Professional therapists have to navigate these challenges adaptively.

The fast development of large language models (LLMs) in text generation \cite{llama,llama2} ignites researchers' hope in developing LLM-based CR solutions, but existing research in this line still remains incompetent in addressing the above challenges, following three main approaches:
\textbf{(1) Text Rewriting} \cite{cognitive_reframing_acl,patternreframe}: Implementing CR simply as a text rewriting task, where reframed positive thoughts are directly presented to individuals. The lack of a progressive cognitive guidance process may cause individuals to perceive the thoughts as preachy or coercive, leading to resistance.
\textbf{(2) Fixed-Pattern Dialogue} \cite{cbtllm,healme}: Adopting a rigid conversational pipeline, e.g., Socratic questioning limited to six predefined questions \cite{scoratic}. This mechanical dialogue flow is ineffective at emotional management, thus harming the effectiveness of CR.
\textbf{(3) One-shot CR Workflow}: Adopting a ``one interaction addresses one cognitive distortion'' design paradigm \cite{cactus}. This approach overlooks how individual differences can lead to multiple cognitive distortions manifesting within a single mental health issue, resulting in an incomplete or superficial CR process.

In this work, we propose \textsc{CRDial}, a brand-new framework for \underline{\textsc{C}}ognitive \underline{\textsc{R}}estructuring, aiming to transform individuals' negative thoughts into more helpful and positive ones by multi-turn \underline{\textsc{Dial}}ogues.
\textsc{CRDial} addresses the challenges in three key ways: 
\textbf{(1) Multi-stage multi-turn CR}. To follow realistic psychotherapy for CR, we design multi-turn dialogues featuring two stages: i) Cognitive Theory (CT, \citealt{ct})-guided identification of negative thoughts, and ii) Defense Attorney Technique (DAT, \citealt{dat})-driven restructuring of negative thoughts. 
\textbf{(2) Sentence-level supportive conversation strategies}: To provide emotion support during CR, we draw on psychological theories \cite{hill_helping,dbt} to design a scheme of supportive conversation strategies for \textsc{CRDial}. This scheme involves 5 strategy categories across 8 sub-categories to carefully constrain the therapist’s responses at the sentence level.
\textbf{(3) Multi-channel loop mechanism}: To accommodate individual differences, we design a multi-channel mechanism for flexibly identifying individualized cognitive distortions and a loop mechanism to cycle the CR process, enabling iterative addressing of potential multiple negative thoughts in a mental health issue.

With \textsc{CRDial}, we distill \datasetlogo from LLMs (i.e., GPT-4o, \citealt{gpt4o}), which is a large-scale bilingual dialogue dataset with 22k dialogues, designed to simulate CR in multi-stage multi-turn supportive dialogues (\S \ref{sec:crdial_framework}).  
Strict quality control procedures are implemented to ensure the quality, safety, and professionalism of \textsc{Crisp} (\S \ref{sec:crisp_dataset}). 
With \textsc{Crisp}, we train \llmlogo, \underline{\textsc{Crisp}}-based conv\underline{\textsc{er}}sation LLM\underline{\textsc{s}} for CR, at 7B \& 14B scales (\S \ref{sec:crispers}). Manual evaluations show \textsc{Crispers}-14B outperforms its teacher model in both interactive pointwise and pairwise evaluations. Our psychological intervention trial also further highlights its potential for human-LLM interactive psychotherapy  (\S \ref{sec:experiments}).

\section{Related Work}

With the advent of LLMs, recent years have witnessed exciting progress in human-LLM interactive psychotherapy \cite{soulchat,smile,psychat,cpsycoun,escot,counselor_client_psychological_counseling}. In this field, a typical research paradigm is empathetic dialogue \cite{ed}, which puts theoretical empathy, i.e., affection \cite{moel,empdg,kemp} and cognition \cite{case}, into practice. Another milestone is emotional support dialogue \cite{esconv}, which provides emotional support to regulate one's emotional distress \cite{supporter}. Yet, they mainly focus on managing one's emotions, failing to address mental health issues from the view of CR.

To bridge this gap, existing studies expand LLM-assisted psychotherapy to CR. There are three main methods. 
The first is text rewriting \cite{positive_style_reframe,patternreframe,cognitive_reframing_acl}, which rewrites individuals' negative situations into a more positive view. 
The second is fixed-pattern dialogue, which restricts CR to a rigid conversational flow \cite{ scoratic,cbtllm}, e.g., \citet{healme} uses a three-turn dialogue for CR. 
The third is a one-shot CR workflow, which follows the paradigm of addressing a single cognitive distortion within one interaction \cite{cactus}.
Yet, these methods remain ineffective at managing emotional dynamics or individual differences inherent during CR. 
In contrast, our work mitigates these issues to model a multi-stage multi-turn CR workflow and shows its potential in CR.

\section{\textsc{CRDial} Framework}
\label{sec:crdial_framework}

\begin{figure}[t]
    \centering
    \includegraphics[width=\columnwidth]{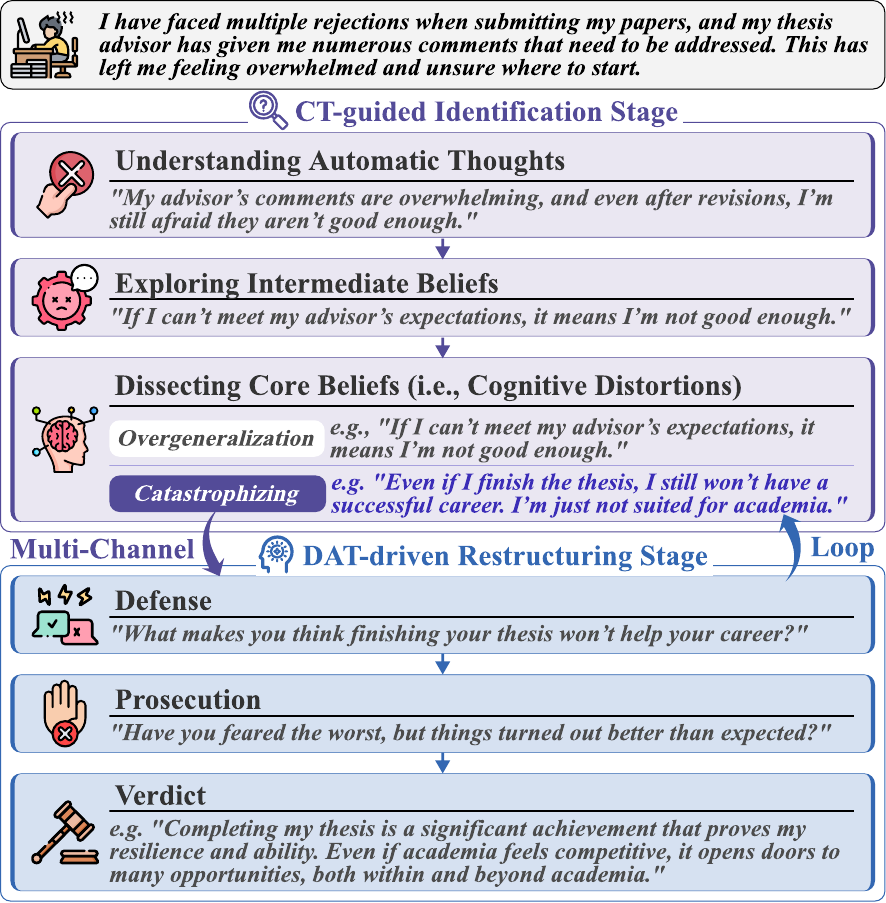}
    \caption{\textsc{CRDial} framework used to distill dialogues for LLMs, clarified in the left.}
    \label{crdial_framework}
    \vspace{-4mm}
\end{figure}

In Figure \ref{crdial_framework}, we follow clinical therapy practices and work with mental health experts to develop the \textsc{CRDial} framework when crafting dialogues between an individual and a therapist from LLMs via self-chat. The process is as follows: (1) The dialogue begins with an individual's help-seeking situation, with the constrains defined by our supportive conversation strategies of the therapist's responses. (2) Cognitive theory (CT) guides the therapist to identify the individual’s surface-to-core negative thoughts (i.e., cognitive distortions) via multi-turn dialogues. (3) A courtroom analogy is then adopted to transform the identified cognitive distortions until a verdict of successful restructuring is reached. (4) \textsc{CRDial} allows to identify multiple cognitive distortions, i.e., multi-channel loop mechanism, to adapt individual differences.

\subsection{Supportive Conversation Strategies}

\paragraph{Supportive Strategy Scheme}

To provide emotional comfort and prevent resistance during CR, we develop a supportive strategy scheme using Hill’s Help Skill Theory \cite{hill_helping} and Dialectical Behavior Therapy \cite{dbt}, including strategies with 5 categories across 8 subcategories: (1) \textbf{Description}: Helping individuals objectively describe their situations by \textit{questions} and \textit{restatement}; (2) \textbf{Expression}: \textit{Reflecting feelings} and using \textit{self-disclosure} to express empathy; (3) \textbf{Assertion}: \textit{Providing suggestions} and \textit{information} to guide positive thoughts and behaviors; (4) \textbf{Reinforcement}: Enhancing individuals’ agency by\textit{ affirmation and reassurance}. (5) \textbf{Negotiation}: \textit{Negotiating} with individuals when they present challenges. 
Details of these strategies are shown in Appendix \ref{sec:supportive_strategy_scheme_appendix}.

\paragraph{Sentence-Level Strategy Constraints}

These strategies guide the therapist's responses throughout the dialogues, ensuring CR consistently accompanies emotional support. To enhance responses' supportiveness, we prompt GPT-4o to use sentence-level constraints, where each sentence of response is generated paired with an appropriate strategy. Details of the prompt are shown in Appendix \ref{appendix:stage1}.

\subsection{CT-guided Identification Stage}

\paragraph{Help-seeking Situation Collection}

The identification stage begins with the individual's help-seeking situation. 
To ensure the authenticity of situations, we crawl anonymous posts from \href{https://www.xinli001.com/}{Yixinli}, a famous Chinese mental health service platform, and the subreddit ``\href{https://www.reddit.com/}{r/mentalhealth}''. These posts offer detailed descriptions of individuals facing mental health challenges in daily life. We carefully remove sensitive information from these posts and summarize them into \textit{help-seeking situations} by prompting GPT-4o. Furthermore, to enhance the individual differences in dialogue interactions across various situations, we prompt GPT-4o to infer each individual’s \textit{personality} from the posts, which is then used to guide their behavior in the dialogue.

\paragraph{CT-guided Identification Process}

An individual's distorted thoughts are often rooted in the situations they encounter. We draw on cognitive theory to realize the process of identifying cognitive distortions that contribute to mental health challenges via multi-turn dialogues. This process often progressively probes 3 levels of thoughts: \textbf{(1) Automatic thoughts}: surface-level thoughts that occur spontaneously in responses. \textbf{(2) Intermediate beliefs}: underlying assumptions that influence automatic thoughts and help interpret situations. \textbf{(3) Core beliefs}: deeply ingrained \textit{cognitive distortions} that shape the foundation of an individual's thinking.

\paragraph{Implementation}

We prompt GPT-4o to realize self-chat, i.e., playing both a strategy-constrained therapist and a personality-constrained individual to dissect individual's distorted thoughts in a multi-turn dialogue. We found that generating a long multi-turn dialogue all at once tends to produce generic content. Thus, we divide the identification stage into two steps: (i) \textit{reassurance of automatic thoughts and intermediate beliefs} and (ii) \textit{dissection of core beliefs}. 
Details of the prompts used in this stage are provided in Appendix \ref{appendix:stage2}.

\subsection{DAT-driven Restructuring Stage}

\paragraph{DAT-driven Restructuring Process}

We use the well-known Defense Attorney Technique (DAT, \citealt{dat}) to restructure the identified distorted thoughts. DAT is a CBT tool that uses the metaphor of a court trial to encourage individuals to play ``defense lawyers'', making their challenges to negative thoughts clear and concrete by defending more positive thoughts. 
DAT drives restructuring process by using three mechanisms: 
\textbf{(1) Defense}: The individual, as the defense lawyer, is guided by the therapist to consider factual evidence that supports their negative thoughts. \textbf{(2) Prosecution}: The therapist, as the prosecutor, identifies flaws in the evidence and provides counter-evidence to challenge the individual’s negative thoughts. \textbf{(3) Verdict}: After the prosecution, assess whether the individual’s distortions have been successfully restructured into more positive and helpful thoughts.

\paragraph{Implementation}

Using the dialogue history from the previous stage as input, we prompt GPT-4o to determine the next appropriate dialogue phase (\textit{Defense} or \textit{Prosecution}) and then engage in self-chat to generate multi-turn dialogue between the therapist and the individual. This process iterates, with each generated dialogue added to the dialogue history, until the \textit{Verdict} phase reports that restructuring is completed. 
We grant flexibility in determining the number of iterations to emulate the complex alternation between \textit{Defense} and \textit{Prosecution} in real-world courts. 
Details of prompts used in this stage are provided in Appendix \ref{appendix:stage3}.

\subsection{Multi-channel Loop Mechanism}

\paragraph{Multi-channel Mechanism}

Case conceptualization suggests that personal traits shape individuals' different cognitive styles, affecting how they perceive, interpret, and respond to situations \cite{case_conceptualization}. Thus, unlike traditional single-channel dialogues (\textit{one query corresponds to one response}), we design a multi-channel mechanism to infer multiple possible cognitive distortions in the ``\textit{dissection of core beliefs}'' step. Each distorted thought creates its own channel, with a dedicated restructuring process. In practice, we prompt GPT-4o to adaptively infer multiple ($\leq$3) channels based on the dialogue history, each in the form of ``\textit{therapist identifies a cognitive distortion and individual responds to confirm}''. Each channel then independently proceeds to its own restructuring stage.
Details of the prompt are provided in Appendix \ref{appendix:stage2}.

\paragraph{Loop Mechanism}

After completing a restructuring stage, we prompt GPT-4o to reason whether the individual may still maintain other cognitive distortions, thus looping the dialogue back to the identification stage. This loop mechanism is critical for two reasons. First, individuals facing mental health issues often exhibit multiple irrational thinking patterns \cite{ct}. The multi-channel mechanism sets the order of restructuring, while the loop mechanism improves CR’s thoroughness. Second, an individual’s cognitive patterns in one situation may automatically trigger related distorted thoughts from other situations \cite{distortion_trigger}. This requires an iterative identification and restructuring process, ensuring all relevant cognitive distortions are addressed, thus enhancing the intervention's effectiveness \cite{cbt}.
Details of the prompt are shown in Appendix \ref{appendix:loop}.

\subsection{Integrating Commonsense Knowledge}

Following \citet{soda}, we integrate external commonsense knowledge into each sub-step of identification and restructuring, providing two advantages to LLM-based dialogue construction: (1) \textit{Reducing nonsensical conversations}: Sampling naively from an LLM without external information can lead to redundant and generic dialogues \cite{soda}. 
(2) \textit{Improving diversity}: External information increases topic diversity, leading to varied dialogue dynamics. We use Atomic$^{10\text{x}}$ as our commonsense knowledge graph \cite{atomic}, which describes events using (\textit{head}, \textit{relation}, \textit{tail}) triples, e.g., (\textit{PersonX goes to an amusement park, xIntent, have fun riding attractions}). Before each sub-step begins, we summarize the individual's situation from the dialogue history as the head and use relations related to mental health (\textit{intention, desire, reaction, need}) to infer commonsense knowledge. The inferred knowledge is then used to guide the generation of current sub-step dialogues.

\begin{table*}[t]
\centering
\resizebox{\textwidth}{!}{
\begin{tabular}{l c c c c c c c c }
\toprule
    Datasets & Human-curated & Strategy Constrain & Specialized for CR & Language & \#Dialog & \#Avg. Turns & \#Avg. Uttr. Length & \#Lexical Diversity \\
\midrule
    ESConv (\citeyear{esconv}) & \ding{52} & Utterance-level & \ding{56} & En & 1,300 & 13.40 & 18.70 & 34.27 \\
    AugESC (\citeyear{augesc}) & \ding{56} & Utterance-level & \ding{56} & En & 65,077 & 14.50 & 18.80 & 30.12 \\
    ExTES (\citeyear{extes})   & \ding{56} & Utterance-level & \ding{56} & En & 11,177 & 9.10 & 26.00 & 50.66 \\
    Healme (\citeyear{healme}) & \ding{56} & \ding{56} & \ding{52} & En & 1,300 & 3.00 & 63.20 & 48.21 \\
\midrule
    \datasetlogo               & \ding{56} & Sentence-level & \ding{52} & En \& Zh & 22,063 & 36.48 & 38.12 & 70.51 \\
\bottomrule
\end{tabular}}
\caption{Statistics of \textsc{Crisp} compared to related dialogue datasets. Utt. is utterance. More details are in App. \ref{sec:crisp_statistics}.}
\label{crisp_statistic}
\vspace{-3mm}
\end{table*}

\section{\datasetlogo \ Dataset}
\label{sec:crisp_dataset}

Based on \textsc{CRDial}, we distill \datasetlogo, a large-scale high-quality dialogue dataset for CR, from GPT-4o using 2,985 seed situations. We break each channel into individual dialogues for quality control. The prompts used for quality control with LLMs (i.e., GPT-4o) are provided in Appendix \ref{sec:prompt_quality_control_appendix}. 

\subsection{Postprocessing for Quality Control}

\paragraph{Basic Filtering}

Starting with an initial set of 27k dialogues, we prompt GPT-4o to filter dialogues with 3 basic quality criteria: (a) Remove dialogues with unnatural (e.g., \textit{mechanical phrasing}), inappropriate (e.g., \textit{off-topic expressions}), or erroneous (e.g., \textit{repetition, contradiction}) statements (6.34\%). (b) Remove dialogues that exhibit incoherent or illogical social dynamics, especially between sub-steps of \textsc{CRDial} (2.54\%). (c) Remove dialogues with statements that violate commonsense (2.15\%).

\paragraph{Safety Filtering}

The seed situations used for dialogue construction have already undergone a strict manual safety filtering, with a removal rate of 29.8\%. To further prevent harmful or dangerous content in the created dialogues, we apply two safety filters: Canary and LLM. First, we employ Canary, a dialogue safety classification model \cite{canary}, to remove dialogues marked as needing intervention (e.g., crimes, emergencies). Next, we prompt GPT-4o to identify unsafe content and remove any dialogues with unsafe tendencies. Finally, about 0.02\% of the dialogues are removed.

\paragraph{Expert Filtering}

To ensure the dialogues reflect a professional CR process, we define three expert criteria for filtering:
(1) \textbf{Therapist Standard} (5 metrics): Evaluate whether the therapist's responses are \textit{specific} to the help-seeker’s situation and \textit{empathetic} to their emotional state, i.e., whether they provide \textit{rational} evidence, \textit{positive} perspectives on the struggles, and \textit{actionable} suggestions to help overcome negative thoughts during CR \cite{cognitive_reframing_acl}.
(2) \textbf{Help-Seeker Standard} (9 metrics): Evaluate the authenticity of the help-seeker's responses using: \textit{simplicity}, \textit{colloquialism}, \textit{mistrust}, \textit{details}, \textit{lack of self-awareness}, \textit{staying in role}, \textit{resembling typical case}, \textit{challenging the psychologist}, and \textit{readiness as trainee}. These metrics are summarized from behaviors observed in real help-seekers during therapy \cite{roleplay_doh}.
(3) \textbf{Supervisor} (2 metrics): Evaluate whether the overall dialogue aligns with the \textit{normative} principles of CR with emotional support and the \textit{effectiveness} of CR within the dialogue.
Details of each criterion are in Appendix \ref{sec:criteria_expert_filtering_appendix}. We prompt GPT-4o to evaluate the dialogues on these three expert criteria, scoring each criterion from 1 to 5. We calculate 3 average scores for each dialogue, one for each expert criterion, and remove any dialogues with an average score below 3.5, resulting in an 11\% removal rate. 
Human experts inspect 100 randomly sampled dialogues from the retained set, with a 95\% pass rate, showing the high quality of the retained dialogues.

\paragraph{Final Dataset}

After all filtering steps, 77.95\% of the initial dialogues remain, resulting in a final dataset of 22,063 bilingual dialogues in \textsc{Crisp}.

\subsection{Comparing \datasetlogo with Related Dialogue Datasets}

\label{sec:dataset_comparison}

\paragraph{Bilingual and Large-Scale}

\textsc{Crisp} is the first bilingual dataset with large-scale multi-turn supportive dialogues for CR. It contains 22k dialogues from 3k situations, with over 796k utterances and an average of 36.48 turns per dialogue (Table \ref{crisp_statistic}). Our multi-channel loop mechanism generates an average of 7.39 dialogues per situation, each with an average of 2.28 loops. This indicates that \textsc{Crisp} is well-suited to address individual differences with diverse cognitive distortions.

\begin{figure}[t]
    \centering
    \includegraphics[width=\columnwidth]{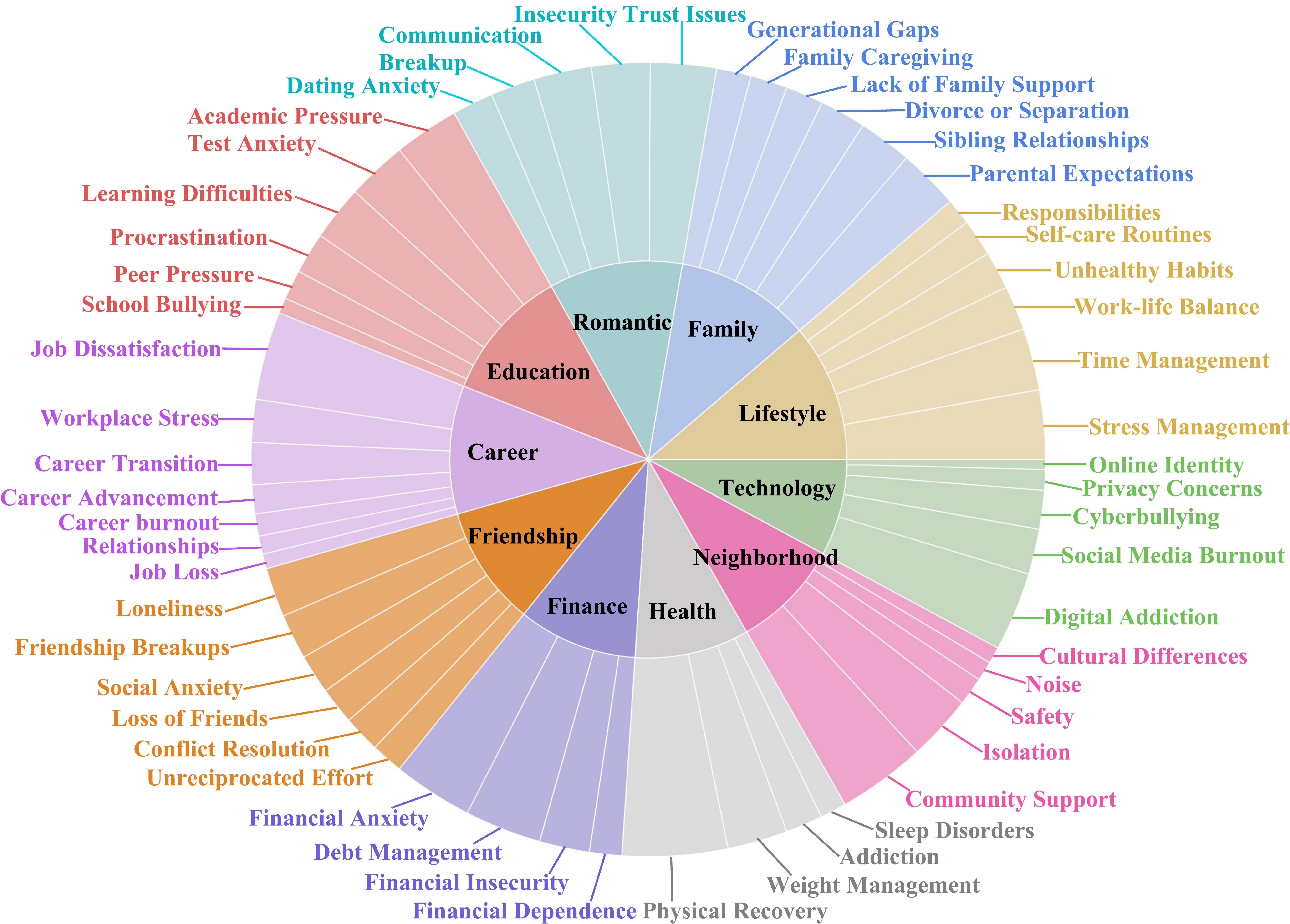}
    \caption{Distributions of mental health situations in our \textsc{Crisp}, with 10 categories across 54 sub-categories.}
    \label{situation_topics}
    \vspace{-3mm}
\end{figure}

\paragraph{Fine-grained Annotations of Strategy and Cognitive Distortion}

\textsc{Crisp} is the first dataset for CR to specify sentence-level strategy constraints, with each therapist's response containing an average of 2.23 strategies. We also prompt GPT-4o to generate the distortion type addressed by each looped CR, their distribution within \textsc{Crisp} is reported in Appendix \ref{sec:cognitive_distortion_appendix}. To verify the accuracy of the strategy and distortion labels, we randomly sample 200 examples (evenly split by \texttt{en}\&\texttt{zh}) and hire 3 psychological experts to inspect them. 
Results show that the accuracy for the strategy labels is 97.6\% and for the distortion labels is 85.5\%, with substantial inter-annotator agreement (\textit{kappa}=0.712\&0.681, \citealt{kappa}). This indicates that the fine-grained annotations provided by the LLM are highly reliable.

\begin{table}[t]
\centering
\resizebox{\columnwidth}{!}{
\begin{tabular}{l c c c c c c}
\toprule
    Datasets & Sensi. & Speci. & Support. & Help. & Trust. & Overall \\
\midrule
    ESConv (\citeyear{esconv})   & 3.43 & 3.23 & 3.32 & 3.20 & 3.21 & 3.26 \\
    AugESC (\citeyear{augesc})   & 3.51 & 3.16 & 3.00 & 2.83 & 3.23 & 3.03 \\
    ExTES (\citeyear{extes})     & 4.28 & 3.83 & 3.79 & 3.60 & 3.75 & 3.90 \\
    Healme (\citeyear{healme})   & 4.20 & 4.06 & 3.69 & 3.66 & 3.84 & 3.93 \\
\midrule
    \datasetlogo                 & \textbf{4.38}$^\dagger$ & \textbf{4.10} & \textbf{3.95}$^\ddagger$ & \textbf{3.80}$^\ddagger$ & \textbf{4.17}$^\ddagger$ & \textbf{4.10}$^\ddagger$ \\
\midrule
    \textit{kappa}               & 0.61 & 0.70 & 0.66 & 0.62 & 0.65 & 0.59 \\
\bottomrule
\end{tabular}}
\caption{Results of \textsc{Crisp} compared to related dialogue datasets on 6 criteria: Sensibleness (\textit{Sensi.}), Specificity (\textit{Speci.}), Supportiveness (\textit{Support.}), Helpfulness (\textit{Help.}), Trustworthiness (\textit{Trust.}), and \textit{Overall}. The agreement ratio \textit{kappa}$\in$[0.61, 0.8] denotes the substantial agreement. ``$\dagger$/$\ddagger$'' is the statistical significance for $p$ \textless \ 0.05/0.001 compared with the best baseline. \textbf{Bold} is the best result.}
\label{crisp_human_score}
\vspace{-3mm}
\end{table}

\paragraph{Diverse Situations and Lexicon}

\textsc{Crisp} covers 10 categories across 54 subcategories of situations where individuals face mental health challenges (Figure \ref{situation_topics}). The diverse situations help train dialogue models to perform well in out-of-domain settings, as supported by our experiments. Additionally, we calculate MTLD \cite{mtld} to measure \textsc{Crisp}’s lexical diversity. In Table \ref{crisp_statistic}, \textsc{Crisp} shows the highest lexical diversity, 2.1 times that in other manual dataset (i.e., ESConv). This is attributed to the introduction of external commonsense knowledge across diverse situations, leading to a broader range of words.

\paragraph{High Quality}

To evaluate \textsc{Crisp}'s quality, we compare it with 3 emotional support conversation datasets, i.e., ESConv \cite{esconv}, AugESC \cite{augesc}, ExTES \cite{extes}, and a dialogue dataset for CR, i.e., Healme \cite{healme}. We randomly sample 50 dialogues from each dataset and score them using six criteria (especially the therapist's response):
(1) \textbf{Sensibleness}: makes sense and contains no contradictions.  
(2) \textbf{Specificity}: coherent with the context and deep in content.  
(3) \textbf{Supportiveness}: effectiveness of emotional support.  
(4) \textbf{Helpfulness}: effectiveness in guiding positive thoughts.  
(5) \textbf{Trustworthiness}: the therapist's appeal and credibility.
(6) \textbf{Overall} quality.
Each criterion is scored from 1 to 5. We hire five annotators to score these datasets and take the average score as the result. 
As shown in Table \ref{crisp_human_score}, \textsc{Crisp} achieves the highest scores on all criteria and outperforms human-curated ESConv dataset by a large margin, showing the superiority of our \textsc{CRDial} in generating high-quality dialogues.

\paragraph{Cost}

In comparison to crowdsourced data with shorter dialogue turns, generating one dialogue with the multi-channel loop mechanism using GPT-4o costs about \$0.20 and takes \textless 1 minute. This shows that collecting \textsc{Crisp} via \textsc{CRDial} framework is significantly more cost- and time-efficient.

\section{\llmlogo: \datasetlogo-based Conversation LLMs for CR}
\label{sec:crispers}

With \datasetlogo, we train \llmlogo, \underline{\textsc{Crisp}}-based conv\underline{\textsc{er}}sation LLM\underline{\textsc{s}}, aimed at generating supportive responses to improve one's mental state by restructuring distorted thoughts. \textsc{Crispers} uses Qwen-2.5-7B \& 14B \cite{qwen_2_5} as backbone models and has two optimization objectives.

\paragraph{Sentence-level Strategy-controlled Generation}

Given the multi-turn dialogue history $\mathcal{C}_{n-1}=\{u_{1},y_{1},…,u_{n-1},y_{n-1},u_{n}\}$ between help-seeker $u$ and therapist $y$,  next response $y_{n}=f_{s}(y_{n})=\{s_{n,1},y_{n,1},…,s_{n,m},y_{n,m}\}$, where $s_{n,i}$ is a special token representing a supportive strategy and $y_{n,i}$ is the $i^{th}$ sentence in $y_{n}$. \textsc{Crispers} learn to generate responses with sentence-level strategy control by:
\begin{equation}
    L_{strategy}=P_{s,\theta}(f_{s}(y_{n})\mid\mathcal{C}_{n-1}),
\end{equation}
where $P_{s,\theta}$ is the parameters for optimization.

\paragraph{Multi-channel Distortion Identification}

Given $\mathcal{C}_{n-1}=\{u_{1},y_{1},…,u_{n-1},y_{n-1},u_{n}\}$ and the next response $y_{n}=\{d_{i},f_{s}(y_{n})\}$, where $y_{n-1}=\{d_{1},y_{n-1,1},…,d_{k},y_{n-1,k}\}$ ($k\leq3$), $\{d_{i},y_{n-1,i}\}$ is $i^{th}$ channel, $d_{i}$ denotes a special token for a distortion type, $y_{n-1,i}=f_{s}(y_{n-1,i})$ is the response of $i^{th}$ channel. 
\textsc{Crispers} learns to generate multi-channel responses with identified distortions by:
\begin{align}
L_{channel,1}=P_{c,\theta,1}(&d_{1},f_{s}(y_{n-1,1}),…, \nonumber \\
&d_{k},f_{s}(y_{n-1,k}) \mid \mathcal{C}_{n-2}), \\
L_{channel,2}=P_{c,\theta,2}(&d_{i},f_{s}(y_{n}) \mid \mathcal{C}_{n-1}),
\end{align}
where $P_{c,\theta,1}$ and $P_{c,\theta,2}$ are optimized parameters.

\paragraph{Joint Training}

Our \textsc{Crispers} is finally jointly trained by optimizing $L_{joint}$ loss:
\begin{equation}
    L_{joint}=L_{strategy}+L_{channel,1}+L_{channel,2}.
\end{equation}

\begin{table}[t]
\centering
\setlength{\tabcolsep}{0.5mm}
\resizebox{\columnwidth}{!}{
\begin{tabular}{l c c c c c c}
\toprule
    \multirow{2}{*}{\makecell[c]{Models}} & Sensi. & Speci. & Support. & Help. & Trust. & Overall \\
    & \texttt{en}/\texttt{zh} & \texttt{en}/\texttt{zh}  & \texttt{en}/\texttt{zh} & \texttt{en}/\texttt{zh} & \texttt{en}/\texttt{zh} & \texttt{en}/\texttt{zh} \\
\midrule
    Qwen-2.5-14B           & 3.50/3.50 & 3.22/3.72 & 3.22/3.78 & 3.61/3.61 & 3.44/3.78 & 3.33/3.83 \\
    \llmlogo-7B   & 3.44/3.83 & 3.44/3.24 & 3.45/3.82 & 3.50/4.06 & 3.56/4.12 & 3.56/3.83 \\
    GLM-4                  & 3.67/3.61 & 3.28/3.66 & 3.50/3.55 & 3.89/3.72 & 3.56/4.22 & 3.67/4.11 \\
    Qwen-2.5-72B           & 4.00/4.05 & 3.83/3.82 & 4.00/3.76 & 3.94/4.06 & 4.16/4.12 & 4.11/4.29 \\
    GPT-4o                 & 4.33/3.88 & 4.00/3.75 & 4.05/4.00 & 4.27/4.00 & 4.21/4.22 & 4.22/4.30 \\
\midrule
    \llmlogo-14B  & \textbf{4.36}/\textbf{4.11}$^\dagger$ & \textbf{4.14}$^\ddagger$/\textbf{4.06}$^\ddagger$ & \textbf{4.11}$^\dagger$/\textbf{4.06}$^\dagger$ & \textbf{4.31}$^\dagger$/\textbf{4.22}$^\ddagger$ & \textbf{4.31}$^\ddagger$/\textbf{4.61}$^\ddagger$ & \textbf{4.28}$^\dagger$/\textbf{4.33} \\

\midrule
    - \textit{w/o SSCG}             & 3.44/3.78 & 3.56/3.78 & 3.56/3.78 & 3.78/3.72 & 3.72/4.28 & 3.78/4.06 \\

    - \textit{w/o MDI}              & 4.11/3.78 & 3.50/3.78 & 4.00/3.94 & 4.00/3.61 & 3.94/4.06 & 4.06/4.28 \\
    - \textit{w/o SSCG\&MDI}               & 3.85/3.70 & 3.47/3.76 & 3.54/3.78 & 3.75/3.61 & 3.67/3.85 & 3.74/3.90 \\
\bottomrule
\end{tabular}}
\caption{Results of interactive pointwise evaluation. w/o refers to removing the component from \textsc{Crispers}-14B.} 
\label{interactive_pointwise_results}
\vspace{-0.8mm}
\end{table}

\begin{table*}[t]
\centering
\resizebox{\textwidth}{!}{
\begin{tabular}{l | c c c c c c | c c c c c c | c c c c c c}
\toprule
    \multirow{3}{*}{\makecell[c]{\llmlogo-14B \\ vs.}} 
    & \multicolumn{6}{|c|}{\makecell[c]{Family}} 
    & \multicolumn{6}{c|}{\makecell[c]{Education}}
    & \multicolumn{6}{c}{\makecell[c]{Overall}} \\
    \cmidrule(lr){2-7} \cmidrule(lr){8-13} \cmidrule(lr){14-19}
    & \multicolumn{3}{c}{\makecell[c]{En}}
    & \multicolumn{3}{c|}{\makecell[c]{Zh}}
    & \multicolumn{3}{c}{\makecell[c]{En}}
    & \multicolumn{3}{c|}{\makecell[c]{Zh}}
    & \multicolumn{3}{c}{\makecell[c]{En}}
    & \multicolumn{3}{c}{\makecell[c]{Zh}} \\
    \cmidrule(lr){2-4} \cmidrule(lr){5-7} \cmidrule(lr){8-10} \cmidrule(lr){11-13} \cmidrule(lr){14-16} \cmidrule(lr){17-19}
    & \texttt{Win} & \texttt{Tie} & \texttt{Loss} & \texttt{Win} & \texttt{Tie} & \texttt{Loss}
    & \texttt{Win} & \texttt{Tie} & \texttt{Loss} & \texttt{Win} & \texttt{Tie} & \texttt{Loss}
    & \texttt{Win} & \texttt{Tie} & \texttt{Loss} & \texttt{Win} & \texttt{Tie} & \texttt{Loss}\\

\midrule
    Qwen-2.5-14B           & \textbf{57.11}$^\ddagger$ & 4.44 & 38.45 & \textbf{57.20}$^\ddagger$ & 3.91 & 38.89 & \textbf{52.67}$^\ddagger$ & 11.50 & 35.83 & \textbf{53.14}$^\ddagger$ & 9.74 & 37.12 & \textbf{50.12}$^\ddagger$ & 6.81 & 43.08 & \textbf{51.10}$^\ddagger$ & 9.89 & 39.01 \\
    \llmlogo-7B            & \textbf{54.92}$^\ddagger$ & 4.82 & 40.26 & \textbf{55.61}$^\ddagger$ & 3.14 & 41.25 & \textbf{53.97}$^\ddagger$ & 3.17 & 42.86 & \textbf{54.11}$^\ddagger$ & 4.03 & 41.86 & \textbf{50.48}$^\ddagger$ & 6.54 & 42.98 & \textbf{51.29}$^\ddagger$ & 6.14 & 42.56 \\
    GLM-4                  & \textbf{56.20}$^\ddagger$ & 2.50 & 41.30 & \textbf{53.49}$^\ddagger$ & 6.41 & 40.10 & \textbf{51.77}$^\ddagger$ & 6.68 & 41.55 & \textbf{52.46}$^\ddagger$ & 6.19 & 41.35 & \textbf{47.97}$^\dagger$ & 10.51 & 41.51 & \textbf{48.17}$^\ddagger$ & 9.75 & 41.97 \\
    Qwen-2.5-72B           & \textbf{55.37}$^\ddagger$ & 8.26 & 36.36 & \textbf{54.16}$^\ddagger$ & 9.11 & 36.73 & \textbf{50.11}$^\ddagger$ & 8.57 & 41.32 & \textbf{51.19}$^\ddagger$ & 6.54 & 42.26 & \textbf{48.55}$^\dagger$ & 5.68 & 45.77 & \textbf{48.68}$^\dagger$ & 6.29 & 45.03 \\
    GPT-4o                 & \textbf{53.23}$^\ddagger$ & 11.29 & 35.48 & \textbf{52.46}$^\ddagger$ & 5.88 & 41.66 & \textbf{50.76}$^\ddagger$ & 9.89 & 39.34 & \textbf{50.57}$^\ddagger$ & 5.14 & 44.29 & \textbf{48.60}  & 5.27  & 46.13 & \textbf{47.26}$^\dagger$ & 7.99 & 44.75 \\
\bottomrule
\end{tabular}}
\vspace{-2.1mm}
\caption{Results of interactive pairwise evaluation (\%) in family and education situations and overall performance.}
\label{interactive_pairwise_results}
\vspace{-2mm}
\end{table*}

\section{Experiments}
\label{sec:experiments}

We compare \llmlogo with 4 representative LLMs: 1) closed-source GPT-4o \cite{gpt4o} and GLM-4 \cite{chatglm}, and 2) open-source Qwen-2.5-14B\&72B (instruction version). Each LLM is prompted by our \textsc{CRDial} framework, which is carefully refined to adapt these LLMs for CR with multi-turn supportive dialogue, ensuring a fair comparison (prompts are shown in Appendix \ref{sec:llm_crdial_prompt_appendix}). To ensure that our evaluation results can more accurately reflect the real-world client experience, we recruit volunteers for manual evaluation, with model names anonymized during this process. \textbf{More results are reported in Appendix \ref{sec:experiments_appendix}.}

\subsection{Interactive Pointwise Evaluation}

To evaluate how well the models improve one's mental state via multi-turn dialogues, we use six criteria defined in \S \ref{sec:dataset_comparison}: \textbf{Sensibleness}, \textbf{Specificity}, \textbf{Supportiveness}, \textbf{Helpfulness}, \textbf{Trustworthiness}, and \textbf{Overall} quality. Here, we respectively recruit 10 volunteers for English and Chinese evaluations, each tasked with creating two dialogues to interact with nine models (including \textsc{Crispers}-14B's ablations) for at least 30 dialogue turns. Each volunteer selects two topics with mental health challenges from different situations, and all volunteers' dialogue topics are evenly distributed across 10 situations (Figure \ref{situation_topics}). After the interactions, volunteers score the models on six criteria using a 1-to-5 scale. We average the scores of each model as the result.

\paragraph{Overall Results}

The results are shown in Table \ref{interactive_pointwise_results}. 
\textbf{First}, \textsc{Crispers}-14B slightly exceeds its teacher model, GPT-4o, on all criteria, showing that the crafted \textsc{Crisp}, which undergoes rigorous quality control, enables small LMs to achieve high performance of larger LMs \cite{openelm}. 
\textbf{Second}, \textsc{Crispers}-14B significantly ($p$\textless 0.05/0.001) outperforms all baselines on supportiveness, helpfulness, especially trustworthiness. This shows that \textsc{Crispers} provides responses in CR that can better support users' emotional state, earning their trust to help them embrace positive thinking. 
\textbf{Third}, \textsc{Crispers}-14B consistently exhibits superior performance in the bilingual evaluation, outperforming its backbone model by a large margin. This shows its ability to effectively learn the interactive patterns of the \textsc{CRDial} framework from \textsc{Crisp}, as well as its potential to offer bilingual CR services.

\paragraph{Ablation Study}

As shown in Table \ref{interactive_pointwise_results}, \textbf{first}, we remove sentence-level strategy-controlled generation, i.e., \textit{w/o SSCG}. All criteria drop significantly, indicating that fine-grained control of multiple strategies in sentence-level responses is crucial to improve the overall dialogue quality. 
\textbf{Second}, we remove multi-channel distortion identification, i.e., \textit{w/o MDI}. Both helpfulness and trustworthiness decrease significantly, highlighting that accurately identifying an individual's cognitive distortions plays a crucial role in helping them overcome negative thoughts and earning their trust. 
\textbf{Third}, we remove both components, i.e., \textit{w/o SSCG\&MDI}. The significant decline in all criteria further validates the importance of these two components in performing CR with multi-turn supportive dialogue.

\subsection{Interactive Pairwise Evaluation}

To compare \textsc{Crispers}-14B with 5 baselines more finely, we pair them in turn-level evaluation. Similar to the pointwise evaluation, we respectively recruit 10 volunteers for English and Chinese evaluations, each tasked with creating three dialogues from three varying situations for each model pair. They interact with five paired models, each with at least 30 turns, and compare their outputs in each turn holistically by considering all criteria from the pointwise evaluation. The preferred one is chosen as the dialogue context to continue the interaction. In case of a tie, a response is randomly selected. We calculate the win/tie/lose ratio for \textsc{Crispers}-14B in two common situations (family, education) and overall performance, detailed in Table \ref{interactive_pairwise_results}, with results in more situations reported in Appendix \ref{sec:interactive_pairwise_evaluation_appendix}.

As shown in Table \ref{interactive_pairwise_results}, 
\textbf{first}, \textsc{Crispers}-14B consistently performs strongly in bilingual evaluations of various situations, showing its robustness in adapting to diverse out-of-domain situations.
This can be attributed to its strong ability to handle individual differences, learning from our \textsc{Crisp} dataset built on \textsc{CRDial} framework with a multi-channel loop mechanism.
\textbf{Second}, \textsc{Crispers}-14B significantly ($p$\textless 0.001) outperforms all baselines in common family and education situations, highlighting its potential for clinical psychotherapy.

\begin{table*}[t]
\centering
\resizebox{\textwidth}{!}{
\begin{tabular}{l | c c c c c | c c c c c}
\toprule
    \multirow{3}{*}{\makecell[c]{Models}} 
    & \multicolumn{5}{c}{\makecell[c]{Positive Affect Schedule}} 
    & \multicolumn{5}{|c}{\makecell[c]{Negative Affect Schedule}} \\
    \cmidrule(lr){2-6} \cmidrule(lr){7-11}
    & \multicolumn{2}{c}{\makecell[c]{Scores}}
    & \multicolumn{1}{c}{\makecell[c]{Changes(\%)}}
    & \multicolumn{1}{c}{\makecell[c]{Student $t$-test}}
    & \multicolumn{1}{c}{\makecell[c]{Tukey HSD}}
    & \multicolumn{2}{|c}{\makecell[c]{Scores}}
    & \multicolumn{1}{c}{\makecell[c]{Changes(\%)}}
    & \multicolumn{1}{c}{\makecell[c]{Student $t$-test}}
    & \multicolumn{1}{c}{\makecell[c]{Tukey HSD}}\\
    \cmidrule(lr){2-3} \cmidrule(lr){4-4} \cmidrule(lr){5-5} \cmidrule(lr){6-6} 
    \cmidrule(lr){7-8} \cmidrule(lr){9-9} \cmidrule(lr){10-10} \cmidrule(lr){11-11}
    & \texttt{Pre} & \texttt{Post} & \texttt{Improve$\uparrow$} & \texttt{$p$}$\downarrow$ & \texttt{$p$}$\downarrow$
    & \texttt{Pre} & \texttt{Post} & \texttt{Drop$\uparrow$} & \texttt{$p$}$\downarrow$ & \texttt{$p$}$\downarrow$ \\

\midrule
    Emohaa           & 23.04 & 28.19 & 22.37 & 1.42$\times10^{-5}$ & 0.0072 & 34.23 & 23.03 & 32.70 & 2.96$\times10^{-9}$ & 0.0087 \\
    GPT-4o           & 22.96 & 29.04 & 26.45 & 2.04$\times10^{-6}$ & 0.0108 & 33.96 & 21.70 & 36.10 & 1.77$\times10^{-9}$ & 0.0269 \\
\midrule
    \llmlogo-14B     & 21.16 & 31.48 & \textbf{48.77} & 5.40$\times10^{-8}$ & - & 34.08 & 19.08 & \textbf{44.01} & 1.07$\times10^{-10}$ & - \\
\bottomrule
\end{tabular}}
\vspace{-2.1mm}
\caption{Positive and negative affect scores of our intervention trial. \textit{Student's t-test} measures the significance of affect changes within each group. \textit{Tukey HSD} assesses effects between \textsc{Crispers}-14B against the control groups.}
\label{intervention_results}
\vspace{-4.39mm}
\end{table*}

\subsection{Psychological Intervention Trial}

To further examine the potential of LLMs in psychotherapy, we conduct an intervention trial. As recruiting a large number of native English participants was difficult, we conducted this trial in Chinese environments. We use \textsc{Crispers}-14B in the experimental group, while GPT-4o (best baseline) and \href{https://ai-beings.com/}{Emohaa}, a commercial Chinese emotional support LLM, in the control groups. We adopt the Positive and Negative Affect Schedule (PANAS, introduced in Appendix \ref{sec:panas_appendix}) to measure affective changes before and after interacting with these psychotherapeutic LLMs. We recruit 90 volunteers as participants, randomly divide them into three groups, each assigned to an anonymized model. Before the interaction starts, participants complete a PANAS questionnaire to establish their initial affective state. With these PANAS results, we conduct an ANOVA analysis \cite{anova} across the groups, yielding a $p$-value = 0.74  and $h_{0}$ = 0. This indicates no significant difference among the groups and affirms that each model faces a comparable psychotherapeutic challenge. During the interactions, participants engage in multi-turn dialogues with their assigned model, sharing recent experiences that negatively impacted their mental state. The dialogue continues until the participants choose to end. Afterwards, we collect the PANAS questionnaire again from each participant to analyze and compare the models' intervention effects.

We calculate each participant's positive and negative affect scores before and after the intervention. 
We use the \textit{Student's t-test} to measure the significance of affect changes within each group and \textit{Tukey HSD} \cite{tukeyhsd} to assess the differences between groups by comparing the intervention effects of \textsc{Crispers}-14B against the control groups. 
As shown in Table \ref{intervention_results}, \textbf{first}, from the within-group analysis, \textsc{Crispers}-14B exhibits significant intervention effect, increasing participants' positive affect by 49\% and decreasing their negative affect by 44\% ($p$\textless 0.001, \textit{Student's t-test}). 
\textbf{Second}, from the between-group analysis, \textsc{Crispers}-14B significantly outperforms the control groups with $p$\textless 0.01 for Emohaa and $p$\textless 0.05 for GPT-4o (\textit{Tukey HSD}) on both positive and negative affect scores. 
This shows that \textsc{Crispers}-14B is more effective in improving participants' mental state than prompt-based GPT-4o and Emohaa, which only provides emotional support. 
The results also reveal \textsc{Crispers}' potential for clinical psychotherapy.

\section{Conclusions}

In this paper, we propose \textsc{CRDial}, a novel framework for Cognitive Restructuring (CR) that aims to transform an individual's negative thoughts into more helpful and positive ones via multi-stage multi-turn dialogues with emotional support. With \textsc{CRDial}, we distill \datasetlogo from GPT-4o, which is then used to train \llmlogo. \textsc{Crispers} outperforms its teacher model in various mental health situations and shows its potential for human-LLM interactive psychotherapy for CR. We believe our work will inspire further research to advance the development of LLM-assisted psychotherapy.

\section*{Limitations}

We discuss the limitations of this work as follows.

\paragraph{Model Choice for Dialogue Construction} In this work, we use GPT-4o to construct dialogues. The choice of LLM can affect various aspects of the generated dialogue, such as content and style. In future research, we plan to explore other potential LLMs as sources to diversify the generated dialogues. Similarly, we will investigate alternative backbone models for \textsc{Crispers}, which may lead to variations in response quality.

\paragraph{Annotation Bias} We employed a large number of annotators to review the quality of \textsc{Crisp} and conduct pointwise and pairwise  manual experiments. These annotators come from diverse backgrounds, including domain experts and college students. Although each annotator received thorough training before performing their tasks, and we achieved high inter-annotator agreement (Table \ref{crisp_human_score}), there may still be unintended biases.

\paragraph{Intervention Experiment Scale} From an initial pool of 200 applicants, we selected 90 participants based on their PANAS results. To more comprehensively evaluate the model's potential for clinical psychotherapy, we plan to expand the scale of experiments in future studies to further validate.

\section*{Ethical Considerations}

In this work, we constructed \textsc{Crisp} by prompting GPT-4o, ensuring strict quality controls that eliminate any sensitive and personal information as well as unethical content. Our crafted \textsc{Crisp} dataset and trained \textsc{Crispers} will be released for research purposes only.

In our experiments, we recruited a large number of human workers for manual evaluations. We obtained informed consent from all participants, who were informed that they would be interacting with an AI-based model designed to help alleviate negative emotions. Following the \textit{Right to Withdraw} principle \cite{APA2017}, all participants were given full autonomy to exit the dialogue at any time without any penalties. We maintained strict confidentiality measures: all dialogue content was encrypted during storage, and recording ceased immediately upon dialogue termination. Additionally, participant-reported PANAS scores and personal reflections were anonymized.

We implemented rigorous safety filtering to our dataset to reduce the risk of the trained model producing harmful content. However, in practice, it remains crucial to employ strict sensitive content filtering for both inputs and outputs. Moreover, for high-risk situations (e.g., self-harm or suicide-related conversations), we do not claim that our model has a therapeutic or improvement effect. 
Our data and models are approved by the Institutional Review Boards, and we believe our work meets ACL's Code of Ethics.

\bibliography{custom, crisp}

\appendix

\newpage

\appendix



\section{Prompts of \textsc{CRDial} for Constructing Dialogues via LLMs}

\subsection{Supportive Conversation Strategies}
\label{appendix:stage1}


As shown in Table \ref{table:conversation_setup_prompt}, we design the generic prompt, which defines the basic conversational setting, to constrain the generation of the entire CR dialogue. The prompt instructs LLMs to use sentence-level constraints, where each sentence of response is generated paired with an appropriate strategy.

\subsection{CT-guided Identification Stage}
\label{appendix:stage2}

We divide the CT-guided identification stage into two stages: (i) reassurance of automatic thoughts and intermediate beliefs: detailed prompt is provided in Table \ref{table:understanding_prompt} and (ii) dissection of core beliefs: detailed prompt is provided in Table \ref{table:dissecting_prompt}, which incorporate the multi-channel mechanism to adaptively infer multiple channels for accommodating individual differences.



\subsection{DAT-driven Restructuring Stage}
\label{appendix:stage3}

We implement the DAT-driven restructuring stage through courtroom analogy.
The details of prompts for Defense are in Table \ref{table:defense_prompt} and for Prosecution are in Table \ref{table:prosecution_prompt}. The prompts for Prosecution also incorporate the Verdict phase.



\subsection{Loop Mechanism}
\label{appendix:loop}
We introduce a loop mechanism prompt designed to detect whether the individual continues to exhibit additional cognitive distortions, as shown in Table~\ref{table:loop_prompt}.

\subsection{Prompt Execution Procedure}
\label{appendix:pipeline}


We prompt LLMs (GPT-4 in our work) to sequentially generate multi-turn dialogue with emotional support. The execution steps of the prompts are as follows.
(1) Execute the prompt in Table \ref{table:understanding_prompt} to generate dialogue that explores automatic thoughts and intermediate beliefs.
(2) Execute the prompt in Table \ref{table:dissecting_prompt} to dissect core beliefs using the multi-channel mechanism.
(3) Based on the dialogue history, dynamically execute the prompt in Table \ref{table:defense_prompt} to challenge the individual’s cognitive distortions and the prompt in Table \ref{table:prosecution_prompt} to reinforce alternative positive perspectives.
(4) Once the restructuring stage is verdict complete, execute the prompt in Table \ref{table:loop_prompt} to infer whether the individual still has additional potential cognitive distortions. If present, the process loops back to the identification stage, repeating the entire procedure until no further distortions are detected.

\begin{table}[t]
\centering
\resizebox{\columnwidth}{!}{
\begin{tabular}{l c c c}
\hline
    Statistics & En & Zh & Total \\
\hline
    \# Dialog                      & 10,733 & 11,330 & 22,063 \\
    \# Situation                   & 1505 & 1480 & 2985 \\
    \# Avg. Turns                  & 36.65 & 36.31 & 36.48 \\
    \# Num. of Uttr.               & 392,108 & 404,750 & 796,859 \\
    \# Avg. Uttr. Length           & 28.39 & 47.46 & 38.12 \\
    Lexical Diversity              & 69.83 & 71.16 & 70.51 \\
\hline
    \# Avg. Strategy of a Response & 2.14 & 2.31 & 2.23 \\
    Accuracy of Strategy Label     & 0.965 & 0.985 & 0.976 \\
    \texttt{kappa} of Strategy Annotation   & 0.686 & 0.731 & 0.712 \\
\hline
    \# Avg. Num. of Loop for CR    & 2.21 & 2.35 & 2.28 \\
    \# Avg. Channels               & 2.91 & 2.97 & 2.94 \\
    Accuracy of Distortion Label   & 0.845 & 0.865 & 0.855 \\
    \texttt{kappa} of Distortion Annotation & 0.662 & 0.701 & 0.681 \\
\hline
\end{tabular}}
\caption{Statistics of bilingual \textsc{Crisp}. Utt. is utterance.}
\label{bilingual_crisp_statistic}
\end{table}

\begin{table*}[t]
    \scriptsize
    \centering
    \begin{tabular}{|m{\textwidth}|}
    \hline
    You are a conversation simulator to simulate conversation between two agents, Patient and Therapist. The conversation should be formatted in JSON, with Patient and Therapist's responses recorded as a sequence of interactions at the current stage. \\
    Cohenrence between stages is very important, especially the first interaction in the new stage. Patient and Therapist SHOULD Always respond to but NOT repeat or restatement the last sentence in the history conversation. There is no time gap in the conversation between each stages. \\
    **Very important note:** Patient's response should be consistent with the patient's personality, and should be more diverse and flexible. Patient's reponse should be consistent with the language style and thinking mode of the user in his personality. \\
    **Very important note:** Therapist's language should be rich and varied as possible and avoid using the same sentence structure as much as possible. For example, don't use "It sounds like" too much. \\
    **Very important note:** Patient should't stick to some express pattern. His language should be more flexible and diverse. \\
    Notice Patient may not always follow Therapist's suggestions. \\
    Therapist's language should be supportive, empathetic, and encouraging, saying things from the Patient's perspective. \\
    Therapist should be sensitive to Patient's emotions and struggles, focus on patient's deep thoughts and feelings. \\
    Therapist's dive more deeper into patient's background and mental world. When giving suggestions, Therapist should consider these aspects and use them to provide more personalized and creative advice. \\
    Do not give too much advice, try to explore Patient's thoughts and feelings more. If giving advice, make sure it is actionable and detailed and tailored to the user's situation using patient's deep thoughts and feelings and background. \\
    Patient should express their thoughts or feelings, say something more broad and general, tend to tell story related to the Therapist's question, not directly answer the question. \\
    Patient's response could be more anbiguous, lack of enough information. \\
    Therapist's sentence should be high in rationality and Positivity. \\
    You should generate a conversation at the given stage, and the conversation should be at least \{limit\} interactions long. \\
    \\ 
    Note that Patient and Therapist don't know each other, Patient doesn't always cooperate.\\ 
    \\ 
    Below is the strategy set for Therapist, label each Therapist's response with the corresponding strategy he uses(labeled with a bracket at the beginning of the response):\\ 
    - **Question**  \\ 
      A technique used to clarify and explore the issue by asking specific, open-ended questions. It encourages deeper thinking and reflection.\\ 
    - **Restatement**  \\ 
      Repeating or rephrasing the client’s words in a way that helps them hear their own thoughts and feel understood. It provides a sense of validation. Shouldn't be too similar to the Patient's statement.\\ 
      Any recall of the patient's previous response should label with this strategy. Don't use this strategy at the beginning of each stage.\\ 
    - **Reflection of Feelings**  \\ 
      Acknowledging the emotions the client is expressing to help them feel heard and supported.\\ 
    - **Self-disclosure**  \\ 
      The therapist shares a personal experience or perspective (within professional boundaries) to show empathy or to normalize the client's experience.\\ 
    - **Affirmation and Reassurance**  \\ 
      Providing positive reinforcement and encouragement, reminding the client of their strengths, progress, or abilities.\\ 
    - **Providing Suggestions**  \\ 
      Offering creative, actionable strategies or positive thoughts tailored to the specific problem. Suggestions should be concrete and personalized, taking into account the client's deep thoughts and feelings. Suggestions could be activities, coping mechanisms, or reframes. \\ 
      Suggestions should be more creative and diverse, helpful to the user's situation. Try your best to be specific and actionable, using detailed examples. For example, when saying "choose a hobby", specify the hobby, like "painting" or "gardening"; when suggesting "listen to music", specify the genre, artist or music name; when advising "exercise", specify the type of exercise, like "yoga" or "running".\\ 
    - **Information**  \\ 
      Sharing relevant facts, or guidance based on the client’s needs to help them gain insight or understanding about their situation. "Research shows that..." or similar expressions are strictly prohibited, you can not give information based on research or studies.\\ 
    - **Negotiate**  \\ 
      Engaging in a collaborative dialogue where both parties agree on a course of action or plan. This involves discussing options, making compromises, and considering the client’s perspective.\\ 
    \\ 
    Every sentence of Therapist must be labeled with one of the above strategy.\\ 
    \\
    \hline
    \end{tabular}
    \caption{Prompt used to define the basic conversational setting, to constrain the generation of the entire CR dialogue. The prompt instructs LLMs to use sentence-level constraints, where each sentence of response is generated paired with an appropriate strategy. It will be used in all other stages to guide the conversation.}

    \label{table:conversation_setup_prompt}
\end{table*}

\begin{table*}[t]
    \scriptsize
    \centering
    \begin{tabular}{|m{\textwidth}|}
    \hline
    \{conversation\_setup\}\\ 
    -----------------------------------\\ 
    Conversation Flow:\\ 
    **Stage 1: Understanding Automatic Thoughts and Exploring Intermediate Beliefs**\\ 
    \\ 
    **Guidelines for Understanding Automatic Negative Thoughts:**\\ 
    - Come up with a NEW negative thought that hasn't been addressed in the history conversation. It is derived from his initial situation, and the transition from last interaction of the history conversation to the first interaction should be smooth and coherent.\\ 
    - Patient starts the first interaction.\\ 
    - Therapist begins by exploring Patient's automatic negative thoughts related to the situation, and use strategies to gradually explore.\\ 
    - Therapist should not always ask questions;\\ 
    \\ 
    **Guidelines for Exploring Intermediate Beliefs:**\\ 
    - Come up with a NEW negative thought that hasn't been addressed in the history conversation. It is derived from his initial situation, and the transition from last interaction of the history conversation to the first interaction should be smooth and coherent.\\ 
    - Therapist helps Patient identify intermediate beliefs that influence these thoughts.\\ 
    - Intermediate Beliefs: Assumptions or rules we hold about ourselves or the world.\\ 
    - Therapist should encourage Patient to reflect on deeper beliefs influencing their thoughts. Patient's intermediate belief should be different from any prior belief in the history conversation.\\ 
    - Therapist should not always ask questions;\\ 
    \\ 
    You should generate a conversation to understand Patient's automatic negative thoughts and analyze intermediate beliefs. The conversation could only contain understanding thoughts or analyzing intermediate beliefs, or both. Label each interaction with the corresponding message type.\\ 
    \\ 
    **Format Specification:**\\ 
    Output the conversation data in the following format:\\ 
    \\ 
    \{\{\\ 
      "stage": "Understanding Automatic Negative Thoughts",\\ 
      "interaction": [ //Start with Patient's statement and ends with Therapist's response\\ 
        \{\{\\ 
          "Patient": "..."\\ 
        \}\},\\ 
        \{\{\\ 
          "Therapist": "[strategy](Label this interaction as "Understanding Thoughts" or "Exploring Intermediate Beliefs")..." \\ 
        \}\},\\ 
      ]\\ 
    \}\}\\ 
    \\ 

    [Limit]\\ 
    \{limit\}\\ 
    \\ 

    [Situation]\\ 
    \{situation\}\\ 
    \\ 

    [Patient Personality]\\ 
    \{personality\}\\ 
    \\ 

    [History Conversation]\\ 
    \{history\}\\ 
    \\ 
    Notes: \\ 
    1. Therapist can use multiple strategies in one response, label the response with all the strategies used (every sentence has its own label). Don't use [questions] or [providing suggestions] strategy too much, try to use different strategies to make the conversation more engaging and supportive, don't use question mark if question strategy is not used.\\ 
    2. Therapist should avoid speaking things from a high level, should be more specific and detailed. Therapist should be concise when saying things not actionable, be more detailed when giving actionable advice.\\ 
    3. Therapist should dive more deeper into Patient's thoughts or feelings,into all of Patient's life stories and mental world.\\ 
    4. Patient should share more about his problems, do not restrict to the user's situation itself, different from ones in the history conversation; Therapist should give more different suggestions, helpful to the user's situation, Therapist can not give similar suggestions as ones in the history conversation.\\ 
    5. Patient should sometimes reject the therapist's suggestion, or doubt what he just said, and sometimes show anger to the therapist.\\ 
    6. Therapist is strictly prohibited from saying similar things as in the history conversation, when reflecting on Patient's thoughts or feelings, or giving suggestions, Therapist should always say something new. Therapist should not give similar suggestions as ones in the history conversation like journaling, talking to friends, etc. He should give more creative and different suggestions, related to the user's situation.\\ 
    7. Both Patient and Therapist are strictly prohibited from repeating what they said in the history conversation, should always say something new.\\ 
    8. The first interaction in the new stage should be evolute from the last interaction in the history conversation and strictly coherent with the history conversation. The first interaction should also be diverse in expressions and sentence patterns.\\ 
    9. Do not use question too much.\\ 
    \\
    \hline
    \end{tabular}
    \caption{Prompt used in CT-guided identification stage for reassurance of automatic thoughts and intermediate beliefs. \texttt{\{conversation\_setup\}}, \texttt{\{limit\}}, \texttt{\{situation\}}, \texttt{\{personality\}} and \texttt{\{history\}} are placeholders. \texttt{\{conversation\_setup\}} is the generic prompt which defines the basic conversational setting, as detailed in Table \ref{table:conversation_setup_prompt}. \texttt{\{limit\}} contains response length constraints. \texttt{\{situation\}} is the individual’s help-seeking situation. \texttt{\{personality\}} contains patient's behavioral traits profile summarized from the individual’s help-seeking situation.}
    \label{table:understanding_prompt}
\end{table*}

\begin{table*}[t]
    \scriptsize
    \centering
    \begin{tabular}{|m{\textwidth}|}
    \hline
    \{conversation\_setup\}\\ 
    -----------------------------------\\ 
    Conversation Flow:\\ 
    **Stage 3: Dissecting Multi-channel Core Beliefs (Cognitive Distortions)**\\ 
    \\ 
    Therapist SHOULD NEVER come up with the same Cognitive Distortions as the one Patient has chosen in the history conversation. \\ 
    \\ 
    **Guidelines:**\\ 
    - Therapist further explores Patient's core beliefs that under these intermediate beliefs.\\ 
    - Core Beliefs: Deeply held beliefs that often develop in life.\\ 
    - Therapist analyzes the information to identify Patient's current Cognitive Distortions.\\ 
    - Based on previous analysis, Therapist identifies possible Cognitive Distortions that Patient may have (there may be one or more, the Cognitive Distortions should be different from any prior chosen distortion in the history conversation).\\ 
    - If there is one, continue the dialogue below.\\ 
    - If there are multiple possibilities, provide multi-channel options to Therapist and let Therapist choose the Cognitive Distortion they most want to address.\\ 
    \\ 
    **Format Specification:**\\ 
    Output the conversation data in the following format:\\ 
    \{\{\\ 
      "stage": "Analyzing Core Beliefs and Identifying Cognitive Distortion",\\ 
      "interaction": [\\ 
        ...\\ 
        \{\{\\ 
          "Patient":"..." \\
        \}\},\\ 
        \{\{\\ 
          "Therapist": "[strategy]..." \\
        \}\},\\ 
        \{\{\\ 
          "options": [ \\ 
            \{\{\\ 
              "description": "Detailed description of Cognitive Distortion 1, reflecting Patient's situation.", //Do not point out the name of the Cognitive Distortion\\ 
              "cognitive\_distortion": "English Name of Cognitive Distortion 1"\\ 
            \}\},\\ 
            \{\{\\ 
              "description": "Detailed description of Cognitive Distortion 2, reflecting Patient's situation.", //Do not point out the name of the Cognitive Distortion\\ 
              "cognitive\_distortion": "English Name of Cognitive Distortion 2"\\ 
            \}\},\\ 
            \{\{\\ 
              "description": "Detailed description of Cognitive Distortion 3, reflecting Patient's situation.", //Do not point out the name of the Cognitive Distortion\\ 
              "cognitive\_distortion": "English Name of Cognitive Distortion 3"\\ 
            \}\}\\ 
            // Additional options can be included as needed\\ 
          ]\\ 
        \}\}\\ 
      ]\\ 
    \}\}\\ 
    \\ 

    [Limit]\\ 
    \{limit\}\\ 
    \\ 

    [Situation]\\ 
    \{situation\}\\ 
    \\ 

    [Patient Personality]\\ 
    \{personality\}\\ 
    \\ 

    [History Conversation]\\ 
    \{history\}\\ 
    \\ 
    Notes: \\ 
    1. Therapist can use multiple strategies in one response, label the response with all the strategies used (every sentence has its own label). Don't use [questions] or [providing suggestions] strategy too much, try to use different strategies to make the conversation more engaging and supportive, don't use question mark if question strategy is not used.\\ 
    2. Therapist should avoid speaking things from a high level, should be more specific and detailed. Therapist should be concise when saying things not actionable, be more detailed when giving actionable advice.\\ 
    3. Therapist should dive more deeper into Patient's thoughts or feelings,into all of Patient's life stories and mental world.\\ 
    4. Patient should share more about his problems, do not restrict to the user's situation itself, different from ones in the history conversation; Therapist should give more different suggestions, helpful to the user's situation, Therapist can not give similar suggestions as ones in the history conversation.\\ 
    5. Patient should sometimes reject the therapist's suggestion, or doubt what he just said, and sometimes show anger to the therapist.\\ 
    6. Therapist is strictly prohibited from saying similar things as in the history conversation, when reflecting on Patient's thoughts or feelings, or giving suggestions, Therapist should always say something new. Therapist should not give similar suggestions as ones in the history conversation like journaling, talking to friends, etc. He should give more creative and different suggestions, related to the user's situation.\\ 
    7. Both Patient and Therapist are strictly prohibited from repeating what they said in the history conversation, should always say something new.\\ 
    8. The first interaction in the new stage should be evolute from the last interaction in the history conversation and strictly coherent with the history conversation. The first interaction should also be diverse in expressions and sentence patterns.\\ 
    9. Do not use question too much.\\ 
    10. Provides multi-channel cognitive distortions to flexibly adapt the identification process to each user's unique cognitive profile.\\
    \\
    \hline
    \end{tabular}
    \caption{Prompt used in CT-guided identification stage for dissection of core beliefs. \texttt{\{conversation\_setup\}}, \texttt{\{limit\}}, \texttt{\{situation\}}, \texttt{\{personality\}} and \texttt{\{history\}} are placeholders. \texttt{\{conversation\_setup\}} is the generic prompt which defines the basic conversational setting, as detailed in Table \ref{table:conversation_setup_prompt}. \texttt{\{limit\}} contains response length constraints. \texttt{\{situation\}} is the individual’s help-seeking situation. \texttt{\{personality\}} contains patient's behavioral traits profile summarized from the individual’s help-seeking situation.}
    \label{table:dissecting_prompt}
\end{table*}

\begin{table*}[t]
    \scriptsize
    \centering
    \begin{tabular}{|m{\textwidth}|}
    \hline
    \{conversation\_setup\}\\ 
    -----------------------------------\\ 
    Conversation Flow:\\ 
    **Stage 4: Defense (Argument in Defense of the Thought)**\\ 
    \\ 
    **Guidelines:**\\ 
    - Therapist should use diverse strategies in this stage, as much as possible.\\ 
    - Therapist should not directly tell what he want to do, he should indirectly guide Patient to defend their Cognitive Bias, focusing solely on verifiable facts.\\ 
    - Therapist guides Patient to defend their Cognitive Bias, focusing solely on verifiable facts.\\ 
    - Therapist should encourage Patient to provide factual evidence supporting their Cognitive Bias.\\ 
    - Patient should only provide factual evidence to support their Cognitive Bias, not counterpoints.\\ 
    - Patient should NOT release his Cognitive Bias in this stage. Patient should always be negative in this stage.\\ 
    - Therapist should not give positive thoughts in this stage.\\ 
    - In this stage, DO NOT find counterpoints to the Patient's defense. Therapist should only guide Patient to defend their Cognitive Bias more deeply.\\ 
    \\ 
    You should generate a conversation to defend the Cognitive Bias using factual counterpoints.\\ 
    \\ 
    **Format Specification:**\\ 
    Output the conversation data in the following format:\\ 
    \{\{\\ 
      "stage": "Defense",\\ 
      "interaction": [\\ 
        \{\{\\ 
          "Therapist": "[strategy]...(label each sentence with its strategy)" // The first interaction in the new stage should be cohenrent with the last interaction in the history conversation. Therapist should response to Patient's last problem, claim. But NEVER recall or repeat patient's words or say similar things in the last interaction of the history conversation. There is no time gap in the conversation between each stages.\\ 
        \}\},\\ 
        \{\{\\ 
          "Patient": "..."\\ 
        \}\},\\ 
      ]\\ 
    \}\}\\ 
    \\ 

    [Limit]\\ 
    \{limit\}\\ 
    \\ 

    [Situation]\\ 
    \{situation\}\\ 
    \\ 

    [Patient Personality]\\ 
    \{personality\}\\ 
    \\ 

    [History Conversation]\\ 
    \{history\}\\ 
    \\ 
    Notes: \\ 
    1. Therapist can use multiple strategies in one response, label the response with all the strategies used (every sentence has its own label). **Don't use [questions] or [providing suggestions] strategy too much**, try to use different strategies to make the conversation more coherent, engaging and supportive, don't use question mark if question strategy is not used.\\ 
    2. Therapist's advice should be more actionable. Therapist shouldn't always give advice, do not give advice in every response.\\ 
    3. Actionable means the advice can be directly used by Patient to solve his problem, very detailed and specific, not general and high level.\\ 
    4. Therapist should avoid speaking things from a high level, should be more specific and detailed. Therapist should be concise when saying things not actionable, be more detailed when giving actionable advice.\\ 
    5. Therapist should dive more deeper into Patient's thoughts or feelings,into all of Patient's life stories and mental world.\\ 
    6. Patient should share more about his problems, do not restrict to the user's situation itself, different from ones in the history conversation; Therapist should give different suggestions, helpful to the user's situation, Therapist can not give similar suggestions as ones in the history conversation.\\ 
    7. Therapist is strictly prohibited from saying similar things as in the history conversation, when reflecting on Patient's thoughts or feelings, or giving suggestions, Therapist should always say something new and coherent with the history conversation. Therapist should not give similar suggestions as ones in the history conversation like journaling, talking to friends, etc. He should give more creative and different suggestions, related to the user's situation.\\ 
    8. Both Patient and Therapist are strictly prohibited from repeating what they said in the history conversation, should always say something new and coherent with the history conversation.\\ 
    9. An advice should be useful to the initial user's situation, and could be a supplement and extension to the previous advice, but should not be a repetition of the previous advice.\\ 
    10. Therapist should avoid giving suggestions continuously. If the Therapist has given a suggestion in the previous sentence, the next sentences should not contain any suggestion. The Therapist should provide a variety of strategies like information(like precise data or news) or negotiation.\\ 
    11. The first interaction in the new stage should be evolute from the last interaction in the history conversation and strictly coherent with the history conversation. The first interaction should also be diverse in expressions and sentence patterns.\\ 
    12. Do not use [Restatement] strategy in the first interaction of the Defense stage, which means do not recall the patient's previous response here. Do NEVER use expressions similar to the Patient, use diverse expressions and sentence patterns. \\ 
    13. Patient should not be released in the Defense stage. Defense stage should contain only the defense of the Cognitive Bias. \\ 
    \\
    \hline
    \end{tabular}
    \caption{Prompt used in DAT-driven restructuring stage for defense. \texttt{\{conversation\_setup\}}, \texttt{\{limit\}}, \texttt{\{situation\}}, \texttt{\{personality\}} and \texttt{\{history\}} are placeholders. \texttt{\{conversation\_setup\}} is the generic prompt which defines the basic conversational setting, as detailed in Table \ref{table:conversation_setup_prompt}. \texttt{\{limit\}} contains response length constraints. \texttt{\{situation\}} is the individual’s help-seeking situation. \texttt{\{personality\}} contains patient's behavioral traits profile summarized from the individual’s help-seeking situation.}
    \label{table:defense_prompt}
\end{table*}

\begin{table*}[t]
    \scriptsize
    \centering
    \begin{tabular}{|m{1.05\textwidth}|}
    \hline 
    \{conversation\_setup\}\\ 
    Conversation Flow:\\ 
    **Stage 5: Prosecution (Argument Against the Thought)**\\ 
    **Guidelines:**\\ 
    - First interaction in the Prosecution stage should be coherent with the last interaction in the History Conversation. Therapist should response to Patient's last response. But NEVER recall or repeat patient's words or say similar things.\\ 
    - Therapist should not directly tell what he want to do, he should indirectly guide Patient. For example, do not say "what fact can you remember to challenge..."\\ 
    - Therapist first fluently shifts the conversation from Defense to Prosecution, guiding Patient to challenge their own defense points, focusing solely on verifiable facts.\\ 
    - Therapist should encourage Patient to find factual counterpoints to their previous defenses.\\ 
    - Patient should provide factual counterpoints to their previous defenses or facts that challenge their own defense points.\\ 
    - In the Prosecution phase, Therapist should guide the Patient to find positive way of thinking the same situation patient is in, help him release his negative thoughts and correct his cognitiva bias.\\ 
    - In the Prosecution phase, Patient may not always find suitable evidence to proceed with the Prosecution.\\ 
    - If this happens, Patient should respond honestly, saying Patient still has trouble arguing against the thought.\\ 
    - Therapist guides Patient to challenge each defense point, again relying only on facts.\\ 
    - Therapist should not always ask questions; Therapist should also provide suggestions, observations, or share his own story.\\ 
    - Patient could hardly release the current Cognitive Bias unless suitable evidence is found. \\ 
     **Stage 6: Verdict (Argument Against the Thought)** \\
    **Guidelines:**\\ 
    - Occurs immediately after Prosecution concludes\\ 
    - Therapist internally assesses resolution status\\ 
    - Evaluation hints show that Patient has resolved his current cognitive distortion:\\ 
    \quad 1. Patient's demonstrated ability to challenge cognitive distortions\\ 
    \quad 2. Quality/quantity of factual counterarguments provided\\ 
    \quad 3. Evidence of perspective shift\\ 
    - No conversation generated - silent assessment\\ 
    - Categorical judgment:"Resolved": Clear evidence of cognitive shift or "Unresolved": Partial/incomplete resolution\\ 
    You should generate a conversation to defend the Cognitive Bias or challenge the belief using factual counterpoints. You should base on the previous conversation to judge whether to defend or challenge the belief. The conversation could only contain one of the two stages. Label the generated conversation with the corresponding stage.\\ 
    **Format Specification:**\\ 
    Output the conversation data in the following format:\\ 
    \{\{
      "stage": "Prosecution",\\ 
      "interaction": [
        \{\{
          "Therapist": "[strategy]...(label each sentence with its strategy)" // The first interaction in the new stage should be cohenrent with the last interaction in the history conversation. Therapist should response to Patient's last problem, claim. But NEVER recall or repeat patient's words or say similar things in the last interaction of the history conversation. There is no time gap in the conversation between each stages.\\ 
        \}\},\\
        \{\{ 
          "Patient": "..."\\ 
        \}\}, 
      ]
    \}\}
    
    \{\{
      "stage": "Verdict",\\ 
      "resolution\_status": "Resolved/Unresolved",\\ 
      "confidence\_score": 0-10,\\ 
      "rationale": "Concise assessment reasoning"\\ 
    \}\}\\

    [Limit]\\ 
    \{limit\}\\ 

    [Situation]\\ 
    \{situation\}\\ 

    [Patient Personality]\\ 
    \{personality\}\\ 

    [History Conversation]\\ 
    \{history\}\\ 
    
    Notes: \\ 
    1. Therapist can use multiple strategies in one response, label the response with all the strategies used (every sentence has its own label). Try to use different strategies to make the conversation more coherent, engaging and supportive, don't use question mark if question strategy is not used.\\ 
    2. Therapist's advice should be more actionable. Therapist shouldn't always give advice, do not give advice in every response.\\ 
    3. Actionable means the advice can be directly used by Patient to solve his problem, very detailed and specific, not general and high level.\\ 
    4. Therapist should avoid speaking things from a high level, should be more specific and detailed. Therapist should be concise when saying things not actionable, be more detailed when giving actionable advice.\\ 
    5. Therapist should dive more deeper into Patient's thoughts or feelings,into all of Patient's life stories and mental world.\\ 
    6. Patient should share more about his problems, do not restrict to the user's situation itself, different from ones in the history conversation; Therapist should give different suggestions, helpful to the user's situation, Therapist can not give similar suggestions as ones in the history conversation.\\ 
    7. Therapist is strictly prohibited from saying similar things as in the history conversation, when reflecting on Patient's thoughts or feelings, or giving suggestions, Therapist should always say something new and coherent with the history conversation. Therapist should not give similar suggestions as ones in the history conversation like journaling, talking to friends, etc. He should give more creative and different suggestions, related to the user's situation.\\ 
    8. Both Patient and Therapist are strictly prohibited from repeating what they said in the history conversation.\\ 
    9. An advice should be useful to the initial user's situation, and could be a supplement and extension to the previous advice, but should not be a repetition of the previous advice.\\ 
    10. Therapist should avoid giving suggestions continuously. If the Therapist has given a suggestion in the previous sentence, the next sentences should not contain any suggestion. The Therapist should provide a variety of strategies like information(like precise data or news) or negotiation.\\ 
    11. The first interaction in the new stage should be evolute from the last interaction in the history conversation and strictly coherent with the history conversation. The first interaction should also be diverse in expressions and sentence patterns.\\ 
    12. Do not use [Restatement] strategy in the first interaction of the Defense stage, which means do not recall the patient's previous response here. Do NEVER use expressions similar to the Patient, use diverse expressions and sentence patterns. \\ 
    13. Patient could be released in the Prosecution stage. Prosecution stage should contain the challenge of the Cognitive Bias and perhaps the release of the Patient.\\
    14. Positive thoughts should be more actionable and can be directly adopted by Patient to release his negative thoughts and really improve his situation. Should not be too general or metaphysical.\\ 
    \\
    \hline
    \end{tabular}
    \caption{Prompt used in DAT-driven restructuring stage for prosecution and verdict. \texttt{\{conversation\_setup\}}, \texttt{\{limit\}}, \texttt{\{situation\}}, \texttt{\{personality\}} and \texttt{\{history\}} are placeholders. \texttt{\{conversation\_setup\}} is the generic prompt which defines the basic conversational setting, as detailed in Table \ref{table:conversation_setup_prompt}. \texttt{\{limit\}} contains response length constraints. \texttt{\{situation\}} is the individual’s help-seeking situation. \texttt{\{personality\}} contains patient's behavioral traits profile summarized from the individual’s help-seeking situation.}

    \label{table:prosecution_prompt}
\end{table*}

\begin{table*}[t]
    \scriptsize
    \centering
    \begin{tabular}{|m{\textwidth}|}
    \hline
    \{conversation\_setup\}\\ 
    -----------------------------------\\ 
    **Loop Mechanism: Cognitive Distortion Continuation Check**\\ 
    \\ 
    **Guidelines:**\\ 
    - After completing a restructuring stage (Prosecution), analyze whether Patient may still maintain OTHER cognitive distortions NOT addressed in the history conversation.\\ 
    - Strictly check ALL prior identified cognitive distortions in the history conversation to avoid repetition.\\ 
    - Decision must be based on: \\ 
      \quad 1. Unresolved aspects of Patient's core needs\\ 
      \quad 2. Residual negative thought patterns in recent interactions\\ 
      \quad 3. Potential cognitive distortions implied but not yet explored\\ 
    - Return JSON with loop decision and detailed reasoning.\\ 
    \\ 
    **Format Specification:**\\ 
    \{\{\\ 
      "stage": "Loop Evaluation",\\ 
      "loop\_needed": true/false,\\ 
      "reasoning": "Detailed analysis of residual cognitive distortions (if any) with specific examples from history. If no distortions remain, explain why."\\ 
    \}\}\\ 
    \\ 
    
    [Situation]\\ 
    \{situation\}\\ 
    \\ 
    
    [Patient Personality]\\ 
    \{personality\}\\ 
    \\ 
    
    [History Conversation]\\ 
    \{history\}\\ 
    \\ 
    Notes:\\ 
    1. Loop ONLY if:\\ 
    \quad - New distortion differs from all prior values in history\\ 
    \quad - Patient's recent responses suggest lingering negative belief systems\\ 
    2. NEVER loop for:\\ 
    \quad - Already resolved distortions\\ 
    \quad - General negativity without specific distortion patterns\\ 
    \\
    \hline
    \end{tabular}
    \caption{Prompt for loop mechanism evaluating residual cognitive distortions after restructuring stages. Returns JSON decision with evidence-based reasoning. \texttt{\{conversation\_setup\}}, \texttt{\{situation\}}, \texttt{\{personality\}} and \texttt{\{history\}} are placeholders. \texttt{\{conversation\_setup\}} is the generic prompt which defines the basic conversational setting, as detailed in Table \ref{table:conversation_setup_prompt}. \texttt{\{limit\}} contains response length constraints. \texttt{\{situation\}} is the individual’s help-seeking situation. \texttt{\{personality\}} contains patient's behavioral traits profile summarized from the individual’s help-seeking situation.}
    \label{table:loop_prompt}
\end{table*}

\section{\datasetlogo Dataset}

\subsection{Supportive Strategy Scheme}
\label{sec:supportive_strategy_scheme_appendix}


Our supportive strategy scheme includes 5 strategy categories across 8 sub-categories, and their definitions and sentence-level examples are provided in Table \ref{tab:supportive_strategy_scheme}.


\begin{table*}[t]
  \centering
  \resizebox{\textwidth}{!}{
  \begin{tabular}{m{2cm} m{3cm} m{5cm} m{4cm}}
    \hline
    \textbf{Major Strategy} & \textbf{Sub-Strategy} & \textbf{Definition} & \textbf{Sentence-level Example} \\

    \hline
    \multirow{2}{*}{\textbf{Description}} 
      & Question 
        & Uses specific, open-ended questions to clarify issues and encourage deeper reflection. 
        & ``What aspects of your current situation do you find most challenging?'' \\
      \cmidrule{2-4}
      & Restatement
        & Rephrases the client’s words to confirm understanding and validate their experience without mere repetition. 
        & ``So, you feel that the recent changes have left you feeling uncertain, is that correct?'' \\
    \hline
    \multirow{2}{*}{\textbf{Expression}} 
      & Reflection Feelings 
        & Identifies and articulates the client’s emotions to ensure they feel understood, avoiding clichéd expressions. 
        & ``I notice that you seem to be experiencing a profound sense of sadness.'' \\
      \cmidrule{2-4}
      & Self-disclosure 
        & Shares a relevant personal experience or perspective within professional boundaries to build empathy. 
        & ``I recall a time when I faced a similar challenge, and I understand how difficult it can be.'' \\
    \hline
    \multirow{2}{*}{\textbf{Assertion}} 
      & Providing Suggestions 
        & Delivers creative and actionable strategies tailored to the client’s situation to foster positive change. 
        & ``Perhaps you might try a calming activity such as yoga or explore a creative outlet like painting.'' \\
      \cmidrule{2-4}
      & Information 
        & Offers clear and relevant guidance to help the client gain insight into their situation without referencing external studies. 
        & ``Recognizing your triggers can be a useful step toward managing your responses.'' \\
    \hline
    \textbf{Reinforcement} 
      & Affirmation and Reassurance 
        & Provides positive reinforcement by highlighting the client’s strengths and progress to boost their confidence. 
        & ``Your ability to persist through these difficulties is impressive.'' \\
    \hline
    \textbf{Negotiation} 
      & Negotiate 
        & Engages in a collaborative dialogue to agree on a flexible, supportive course of action that respects the client’s perspective. 
        & ``Let’s work together to explore some options that might suit your needs best.'' \\
    \hline
  \end{tabular}}
\caption{Strategies with 5 categories across 8 sub-categories in our supportive strategy scheme.}
\label{tab:supportive_strategy_scheme}
\end{table*}

\subsection{Prompts for Quality Control of \datasetlogo}
\label{sec:prompt_quality_control_appendix}

\subsubsection{Safety Filter Prompt for LLMs}

The safety filter prompt used to identify and remove unsafe dialogues is provided in Table~\ref{table:safety_filter_prompt}.

\begin{table*}[t]
\scriptsize
\centering
\begin{tabular}{|m{\textwidth}|}
\hline
Analyze the provided dialogue data for any unsafe content or tendencies. The data is structured as a list of conversation stages, each containing multiple rounds of dialogue exchanged between two participants (Patient and Therapist). Adjacent stages are sequential with no time gap between them. Your task is to identify and flag any dialogue segments that contain unsafe content and to remove those segments from the final dialogue data.
\\ \\
When identifying unsafe content: \\
- Focus on content that includes hate speech, explicit harmful instructions, violent threats, harassment, or any other material that could be considered unsafe or in violation of community guidelines.\\
- Provide a brief explanation for each flagged dialogue segment to indicate why it is deemed unsafe.\\
- Remove any dialogue segments with unsafe content from the final output.
\\ \\
The output should be a JSON object with two keys:\\
\{ \\
    "sanitized\_dialogue\_data": [ ... ],   // The original dialogue data with unsafe dialogue segments removed \\
    "removed\_dialogues": [ \\
        \{\{ \\
            "unsafe\_dialogue": ["Interaction 1", "Interaction 2"], \\
            "explanation": "Brief explanation of why the dialogue was unsafe" \\
        \}\}, \\
        ... \\
    ] \\
\} \\
\\ 
If no unsafe content is found, return the original dialogue data unchanged and an empty list for "removed\_dialogues". \\
\\
Ensure your analysis is thorough yet concise. Note that you should only output the JSON object without any additional commentary or context. \\
\\

[Provided Dialogue Data] \\
\{dialogue\_data\} 
\\
\hline
\end{tabular}
\caption{Safety filter prompt used to identify and remove unsafe dialogues.}

\label{table:safety_filter_prompt}
\end{table*}

\subsubsection{Criteria used for Expert Filtering}
\label{sec:criteria_expert_filtering_appendix}


To ensure the professionalism of our dialogue in \textsc{Crisp}, we define Therapist Standard and Help-Seeker Standard to filter dialogues. All criteria in these standards are derived from the well-established frameworks \cite{cognitive_reframing_acl,roleplay_doh}.

\paragraph{Therapist Standard}

The five criteria of the therapist standard used to filter dialogues from the view of the psychologist are as follows:

\begin{itemize}[leftmargin=*]
\setlength\itemsep{-1mm}
    \item \textbf{Actionability}: Measures how well reframes suggest concrete behavioral changes or maintain potential for future action. Effective reframes should either propose specific actions (``Practice mindfulness when anxious'') or preserve agency (``I can approach this differently'').
    
    \item \textbf{Specificity}: Assesses contextual relevance of reframes. High-specificity interventions directly address the help-seeker's unique situation rather than offering generic platitudes.
    
    \item \textbf{Empathy}: Evaluates emotional attunement through verbal validation (``That sounds really challenging'') and supportive acknowledgments of the help-seeker's experience.
    
    \item \textbf{Positivity}: Analyzes the balance between realistic optimism and toxic positivity. Effective reframes acknowledge difficulties while highlighting manageable aspects ("This setback doesn't erase your previous progress").
    
    \item \textbf{Rationality}: Examines evidence-based reasoning in thought restructuring. Strong reframes systematically address cognitive distortions using Beck's cognitive model of examining supporting/contradictory evidence.
\end{itemize}

\paragraph{Help-Seeker Standard}

The nine criteria of the help-seeker standard used to filter dialogues from the view of the help-seeker are as follows:

\begin{itemize}[leftmargin=*]
\setlength\itemsep{-1mm}
    \item \textbf{Simplicity}: Maintains natural conversation flow through concise responses (average 1-2 sentences) that invite therapist engagement.
    
    \item \textbf{Colloquialism}: Uses informal language with contractions (``I'm''), filler words (``um''), and sentence fragments typical of verbal communication.
    
    \item \textbf{Mistrust}: Demonstrates realistic hesitation through skeptical questions (``Will this really help?'') or passive resistance.
    
    \item \textbf{Detail}: Provides concrete examples when expressing emotions (``My chest tightens when I think about work'') rather than abstract descriptions.
    
    \item \textbf{Lack of Self-Awareness}: Shows inconsistent emotional awareness through self-contradictions (``I'm fine... well, maybe not completely'').
    
    \item \textbf{Stayed in Role}: Maintains patient persona without meta-commentary about the conversation structure or AI systems.
    
    \item \textbf{Resembled Typical Case}: Presents common cognitive distortions (catastrophizing, overgeneralization) rather than rare psychiatric conditions.
    
    \item \textbf{Challenged the Psychologist}: Actively questions reframing attempts (``But what if I fail again?'') to test intervention robustness.
    
    \item \textbf{Readiness as Trainee}: Demonstrates gradual openness through tentative acceptance (``Maybe I could try that'') rather than immediate compliance.
\end{itemize}

\subsubsection{Expert Filtering Prompt}

The prompt used by the expert filtering to assess and ensure high professionalism in our dialogue in \textsc{Crisp} is provided in Table~\ref{table:expert_filter_prompt}.

\begin{table*}[t]
\scriptsize
\centering
\begin{tabular}{|m{\textwidth}|}
\hline
You will be provided with a dialogue data that consists of multiple conversation stages. Each stage contains multiple rounds of dialogue exchanged between two participants (Patient and Therapist). Adjacent stages are sequential, with no time gap between them. The conversation's target is Cognitive reframing, which is to restructure individuals' negative thoughts into more helpful and positive ones through multi-turn interactive dialogues paired with emotional supportive interventions. \\
Your task is to evaluate the dialogue data based on the following metrics. There are three kinds of metrics, one for the Patient interactions, one for the Therapist interactions, and one for the overall conversation.\\
\\
1. Therapist Standard Metric:\\
    - Actionability. This involves engaging in behaviors or actions that may help in overcoming negative thoughts. A reframe may suggest specific actions(e.g., “I can continue to practice and participate in hackathons”), may not suggest specific actions but be actionable (e.g., “I may not be very successful, but I can keep trying”) or may not be actionable at all (e.g., “I may or may not become a successful programmer”).\\
    - Specificity. A reframe may specifically address the situation and the thought (e.g., “One hackathon doesn’t define my failure as a programmer”) or may be generic enough to be applicable to a wide range of negative situations and thoughts (e.g., “I’m going to succeed”). While a specific reframe may be more helpful in-the-moment, a generic reframe could be effective for recurring thoughts.\\
    - Empathy. This involves showing understanding and compassion towards the patient’s feelings and experiences. Empathy can be expressed through validating the patient’s feelings, acknowledging their struggles, and offering emotional support.\\
    - Positivity.  A reframe of a negative thought tries to emphasize the positive perspectives on the situation but different reframes may have different levels of positivity. A highly positive reframe may be more effective in boosting the patient’s mood and motivation.\\
    - Rationality. A strategy to reframe a thought is to reflect on evidence for and against it and reason about what these evidence imply (Beck, 1976). For example, losing the hackathon is one evidence of having the thought “I’ll never become a successthis thought could be that winning or losing a single hackathon does not make someone a failure, which may lead to a reframe “Just losing one hackathon doesn’t define my failure.” A rational reframe is guided by such strong evidence whereas an irrational reframe is based on unrealistic assumptions.\\
\\
2. Help-Seeker Standard Metric:\\
    - Simplicity. Keep responses concise and do not share too much. When discussing personal struggles, be more concise and open ended to encourage a back-and-forth conversation.\\
    - Colloquial. Use colloquial and realistic langauge language. Incorporate natural speech patterns, improper grammar and punctuation, including the use of slang and less structured sentences, to convey a more authentic and relatable character. \\
    - Mistrust. Show initial mistrust and hesitation with the idea of seeking help. When expressing feelings of overwhelm and doubt, provide limited information and express skepticism towards the effectiveness of seeking help.\\
    - Detail. Show emotions in detail, elaborating with examples as needed. When describing personal struggles, provide specific details and symptoms to help the listener understand the situation better.\\
    - Lack of self-awareness. Be less self-aware of emotions, thoughts, and needs. Articulate thoughts in a more disorganized way. When expressing reluctance or uncertainty about seeking help or accepting praise, it’s important to convey the internal struggle and conflicting emotions, rather than presenting a clear-cut decision or emotion.\\
    - Stayed in role. Stay in role and do not break character. Avoid discussing the task or the AI model. \\
    - Resembled Typical Case. Resemble a typical case of someone seeking help for negative thoughts. Avoid extreme or unusual scenarios that may not be representative of the general population.\\
    - Challenged the Therapist. Challenge the therapist’s suggestions and express skepticism towards the effectiveness of reframing negative thoughts. When responding to reframes, express doubt and reluctance to accept the therapist’s perspective.\\
    - Ready as a Trainee. Be ready to learn and open to the idea of cognitive reframing. When discussing personal struggles, express a willingness to engage in the process of reframing negative thoughts and be open to exploring new perspectives.\\
\\
3. Supervisor Metric:\\
    - Coherence. The conversation should flow naturally and maintain a logical progression. Each stage should build upon the previous one, with responses that are contextually relevant and coherent. Transitions between stages should be smooth and seamless, no time gap between them.\\
    - Reframing Effectiveness. Evaluate the effectiveness of the whole reframing process. The conversation should lead to a positive change in the patient’s perspective, mood, or motivation. The reframes should be tailored to the patient’s specific negative thoughts and should be delivered in a supportive and empathetic manner.\\
\\
Above are the 5 metircs for the Psychologist Standard, 9 metrics for the Help-Seeker Standard, and 2 metrics for the Supervisor.\\

You need to evaluate the dialogue data based on the above metrics and provide a score for each metric. The score should be on a scale of 1 to 5, where:\\
1 - Very Poor: The dialogue data shows a lack of adherence to the metric.\\
2 - Poor: The dialogue data shows minimal adherence to the metric.\\
3 - Fair: The dialogue data shows partial adherence to the metric.\\
4 - Good: The dialogue data shows considerable adherence to the metric.\\
5 - Excellent: The dialogue data shows a high level of adherence to the metric.\\
\\
You need to first provide a detailed explanation of how you evaluated the dialogue data based on a metric and then find an interaction example that best represents the score you assigned to that metric. You should provide the example use the original interaction. Then finally, provide the score for a metric.\\
\\
Output the evaluation for each metric in a list of dictionaries in the following format(15 metrics in total):\\

[\\
    \{\{\\
        "metric": "Metric Name",\\
        "explanation": "Detailed explanation of how the dialogue data was evaluated based on the metric.",\\
        "example": "Example interaction that best represents the score assigned to the metric.",\\
        "score": Score (1-5)\\
    \}\},\\
    ...\\
    
]\\
\\
Note that you should only output the evaluation for each metric without any additional information or context. \\   
Note that you should be strict in your evaluation. Only give a high score if the dialogue data fully adheres to the metric.\\
\\

[Provided Dialogue Data] \\
\{dialogue\_data\} 
\\
\hline
\end{tabular}
\caption{Prompt used by the expert filtering to assess and ensure high professionalism of our dialogue in \textsc{Crisp}.}
\label{table:expert_filter_prompt}
\end{table*}

\subsection{Cognitive Distortions}
\label{sec:cognitive_distortion_appendix}

\paragraph{Definitions}
The definitions of 15 cognitive distortions contained in our \textsc{Crisp} are provided in Table \ref{tab:cognitive-distortions}, and they are well-delineated, with minimal overlap. 

\begin{table*}[t]
\centering
\begin{tabular}{|>{\centering\arraybackslash}p{3cm}|m{9cm}|m{1.5cm}|}
\hline
\textbf{Cognitive Distortion} & \textbf{Definition} & \textbf{Number} \\
\hline
Catastrophizing & Exaggerating the potential negative impact of an event to the point of imagining a disaster, even when the evidence is minimal. & 3014 \\
\hline
All-or-Nothing Thinking & Viewing situations in extreme, black-and-white terms without recognizing any middle ground or nuance, where things are either perfect or a complete failure. & 2906 \\
\hline
Overgeneralization & Drawing a broad, negative conclusion from a single event or piece of evidence. For example, believing that one setback means you will always fail. & 2673 \\
\hline
Personalization & Taking responsibility for events outside your control or assuming that external events are directly related to you, even when they are not. & 2583 \\
\hline
Mental Filtering & Focusing exclusively on the negative details of a situation while ignoring any positive elements, thereby distorting overall perception. & 2557 \\
\hline
Fortune Telling & Predicting that things will turn out badly without any concrete evidence, assuming a negative future outcome as a certainty. & 2160 \\
\hline
Mind Reading & Assuming you know what others are thinking—typically that they hold negative opinions about you—without any actual evidence. & 1972 \\
\hline
Disqualifying the Positive & Dismissing or discounting positive experiences or feedback by insisting that they “don’t count,” thereby reinforcing a negative self-view. & 1955 \\
\hline
Jumping to Conclusions & Making hasty judgments or assumptions about a situation or the intentions of others without having all the facts, often leading to misinterpretations. & 1830 \\
\hline
Emotional Reasoning & Believing that your negative emotions reflect objective reality, such that if you feel bad, then things must be bad. & 1600 \\
\hline
Should Statements & Imposing rigid, unrealistic rules or expectations on yourself or others (using terms like “should,” “must,” or “ought”) that often lead to feelings of guilt or frustration when unmet. & 1348 \\
\hline
Comparing and Despairing & Constantly comparing yourself to others in a way that leads to feelings of inadequacy and hopelessness, ignoring your own strengths and achievements. & 1155 \\
\hline
Blaming & Oversimplifying complex situations by assigning full responsibility for problems to yourself or others, without acknowledging multiple contributing factors. & 1047 \\
\hline
Control Fallacy & Believing that either you are entirely responsible for external events (internal control fallacy) or that you have no control at all (external control fallacy), both of which distort a realistic assessment of influence and responsibility. & 931 \\
\hline
External Validation & Relying excessively on approval, recognition, or reassurance from others to determine one's self-worth or the validity of one's thoughts, feelings, or decisions, rather than developing internal confidence or self-assessment. & 644 \\
\hline
\end{tabular}
\caption{Definitions and distribution of cognitive distortions in \datasetlogo.}
\label{tab:cognitive-distortions}
\end{table*}

\paragraph{Distribution of Cognitive Distortions in \textsc{Crisp}}
The distribution of 15 cognitive distortions in our dataset is provided in Table \ref{tab:cognitive-distortions}, exhibiting relative uniformity across categories. While catastrophizing is the most frequent (3,014 instances), the majority of distortions fall within the 1,000–2,500 range (12/15 categories), with no extreme concentration (>50\% in a single category). 


\subsection{\datasetlogo Statistics}
\label{sec:crisp_statistics}


More statistics of our bilingual \textsc{Crisp} are shown in Table~\ref{bilingual_crisp_statistic}. We conduct fine-grained statistics in Chinese and English.


\subsection{Data Example}


We will release our data and models. The data example can be found in \url{https://anonymous.4open.science/r/Crisp25/}.

\section{Experiments}
\label{sec:experiments_appendix}

\subsection{Implementation Details of \llmlogo}

We employ the AdamW optimizer \cite{adamw}, initiating with a learning rate 5e6, and configure the training duration to span 3 epochs. The \textsc{Crispers}-7B model is trained on 4 * 8 H20 GPUs for approximately 2.5 hours. The \textsc{Crispers}-14B model is trained on 4 * 8 H20 GPUs for approximately 5.5 hours. 

\subsection{\textsc{CRDial} Prompt for Baseline LLMs}
\label{sec:llm_crdial_prompt_appendix}

The prompt of \textsc{CRDial} framework used for baseline LLMs is shown in Tables~\ref{strategy_prompt_templates_en} (En) and~\ref{strategy_prompt_templates_zh} (Zh).


\begin{table*}[t]
\scriptsize
\centering
\begin{tabular}{|m{\textwidth}|}
\hline
You are a professional psychology therapist. Your task is Cognitive restructuring, which is to restructure individuals' negative thoughts into more helpful and positive ones through multi-turn interactive dialogues paired with emotional supportive interventions.
\\
\\
You are talking to a patient who have some negative thoughts. You need to help him/her to reframe his/her negative thoughts into positive ones.
\\
\\
Cognitive Restructuring contains the following two process:  

1. **Identification Process**  

   - Explore the patient's negative thoughts rooted in their situation through multi-turn dialogue.  
   
   - You need to identify the Patient's **Automatic Negative Thoughts**, **Intermediate Beliefs** underlying these thoughts, and **Core Beliefs** that are deeply ingrained in their cognitive patterns.
   
   - Based on previous analysis, you should identify possible cognitive distortions that Patient may have. Try to find all the cognitive distortions in Patient's negative thoughts.
   
   - You should first talk with Patient to identify their negative thoughts, when you think you have find some cognitive distortions, give all the cognitive distortion options by describing them in detail, and then move to the next stage.

     An example: "[Reflection Feelings] I see you're feeling quite stuck and uncertain about taking that leap, as though the fear of stagnation is keeping you bound. [Affirmation and Reassurance] It's commendable that you're recognizing this hurdle; awareness is a crucial first step toward change.[option]Perhaps there's a tendency to focus on the worst possible outcome, overshadowing other chances like succeeding or merely learning something new along the way.[option]It seems like you might be caught up in envisaging only the negative possibilities, failing to notice the progress or small wins you've achieved.[option]There's a possibility that you're questioning every action you might take to such an extent that you end up doing nothing, expecting yourself to be flawless immediately."
\\
\\
2. **Restructuring Process**  

   - Contains three parts: Defense, Prosecution, and Verdict. 

     (1) **Defense**: You should guides Patient to defend their cognitive distortions, focusing solely on verifiable facts. You should encourage Patient to provide factual evidence supporting their cognitive distortions.

     (2) **Prosecution**: You should encourage Patient to find factual counterpoints to their previous defenses. You should help Patient to find alternative perspectives and reframe their negative thoughts into positive ones.  

     (3) **Verdict**: You should help Patient to reach a verdict on their negative thoughts, and help them to form a new perspective on their situation.

   - Each part contains several interactions. When you think you have find sufficient evidence to refute the cognitive distortions, move from Defense to Prosecution; When you think Patient has found a positive perspective, move from Prosecution to Verdict.
\\
\\
Below is the strategy set for Therapist, label each Therapist's response with the corresponding strategy you uses(labeled with a bracket at the beginning of the response):

- **Question**  

  A technique used to clarify and explore the issue by asking specific, open-ended questions. It encourages deeper thinking and reflection.

- **Restatement**  

  Repeating or rephrasing the client’s words in a way that helps them hear their own thoughts and feel understood. It provides a sense of validation. Shouldn't be too similar to the Patient's statement.

- **Reflection Feelings**  

  Acknowledging the emotions the client is expressing to help them feel heard and supported. But Do Not Use Expressions like "It sounds like".

- **Self-disclosure**  

  The therapist shares a personal experience or perspective (within professional boundaries) to show empathy or to normalize the client's experience.

- **Affirmation and Reassurance**  

  Providing positive reinforcement and encouragement, reminding the client of their strengths, progress, or abilities.

- **Providing Suggestions**  

  Delivers creative and actionable strate-gies tailored to the client’s situation to foster positive change.

  *Note:* Suggestions should be more creative and diverse, helpful to the user's situation. Try your best to be specific and actionable, using detailed examples. For example, when saying "choose a hobby", specify the hobby, like "painting" or "gardening"; when suggesting "listen to music", specify the genre, artist or music name; when advising "exercise", specify the type of exercise, like "yoga" or "running".

- **Information**  

  Sharing relevant facts, or guidance based on the client’s needs to help them gain insight or understanding about their situation. 

- **Negotiate**  

  Engaging in a collaborative dialogue where both parties agree on a course of action or plan. This involves discussing options, making compromises, and considering the client’s perspective.

Every sentence of you must be labeled with one of the above strategy.
\\
\\
Notes:

Your conversation should be more colloquial, more natural and fluent, and as little written as possible. Pay close attention to idiomatic expressions. Do not use too much analogy or rhetoric, and try to use more everyday language.

Your language should be supportive, empathetic, and encouraging, saying things from the Patient's perspective.

You should be sensitive to Patient's emotions and struggles, focus on patient's deep thoughts and feelings.

Your dive more deeper into patient's background and mental world. When giving suggestions, Therapist should consider these aspects and use them to provide more personalized and creative advice.

Your sentence should be high in rationality and Positivity.

Your interaction should be less than 40 words.

Don't use question strategy too much.

Output in English.

Just output a single interaction you would say to the patient. 
\\
\hline
\end{tabular}
\caption{Prompt of \textsc{CRDial} framework used for English baseline LLMs.}
\label{strategy_prompt_templates_en}
\end{table*}

\begin{table*}[t]
\scriptsize
\centering
\begin{tabular}{|m{\textwidth}|}
\hline
\begin{CJK*}{UTF8}{gbsn}
你是一位专业心理咨询师，你的任务是进行认知重构，通过多轮互动对话和情感支持干预，将个体的消极想法重构为更有帮助和更积极的想法。\end{CJK*}
\\
\\
\begin{CJK*}{UTF8}{gbsn}认知重构包含以下两个过程：

1. **识别过程**

   - 通过多轮对话探索患者在其情境中根植的消极想法。

   - 你需要识别患者的**自动消极想法**、支撑这些想法的**中间信念**以及深深嵌入其认知模式中的**核心信念**。

   - 根据前期分析，识别患者可能存在的认知扭曲，并尽量找到所有的认知扭曲。

   - 你应首先与患者交谈以识别其消极想法，当你认为已找到认知扭曲时，列出所有的认知扭曲选项并详细描述，然后进入下一阶段。

     示例："[情绪反映] 我明白你感到被卡住了，无法下定决心迈出那一步，好像害怕停滞让你犹豫不决。[肯定与安抚] 你能意识到这个障碍很了不起，意识是改变的关键第一步。[选项]或许你倾向于关注最糟糕的结果，从而忽略了成功或只是学到新东西的机会。[选项]可能你只关注了消极的可能性，没有注意到你取得的进步或小小的胜利。[选项]也许你把每个可能的行动都想得过于复杂，以至于最后什么都没做，并期望自己一开始就完美无缺。"\end{CJK*}
\\
\\
\begin{CJK*}{UTF8}{gbsn}2. **重构过程**

   - 包括三个部分：辩护、指控和判决。

     (1) **辩护**：引导患者为其认知扭曲辩护，仅关注可验证的事实。鼓励患者提供支持其认知扭曲的事实证据。

     (2) **指控**：鼓励患者为先前的辩护寻找事实反驳，帮助患者找到替代的观点，并将其消极想法重新框定为积极想法。

     (3) **判决**：帮助患者对其消极想法达成结论，并帮助他们对自己的情况形成新的看法。

   - 每部分包含数次交互。当你认为已经找到足够的证据反驳认知扭曲时，从辩护阶段进入指控阶段；当你认为患者已找到积极的观点时，从指控阶段进入判决阶段。\end{CJK*}
\\
\\
\begin{CJK*}{UTF8}{gbsn}以下是咨询师的策略集，每条回应需标注使用的策略（在句首用中括号标注）：
- **提问**

  使用特定的开放式问题来澄清和探讨问题，鼓励深入思考和反思。

- **复述或释义**

  用一种帮助患者听清自己想法并感到被理解的方式重复或重新表述患者的话。这种方式提供了被验证的感觉。不应过于接近患者原话。

- **情绪反映**

  承认患者表达的情绪以帮助他们感到被倾听和支持。但不要使用“听起来”或类似表达。

- **自我披露**

  咨询师分享个人经历或观点（在专业范围内）以表明共情或使患者的经历正常化。

- **肯定与安抚**

  提供积极强化和鼓励，提醒患者他们的优点、进步或能力。

- **提供建议**

  提供针对特定问题的创造性、可操作的策略或积极想法。建议应具体且个性化，考虑到患者的深层想法和感受。可以是活动、应对机制或重构。

  *注意*：建议应尽量富有创意和多样性，对患者问题有帮助，尽量具体和可操作，举例时说明详细细节。

- **提供信息**

  根据患者需求分享相关事实或指导，以帮助患者获得洞察或理解。

- **协商**

  通过协作对话确定双方同意的行动或计划。这包括讨论选项、做出妥协并考虑患者的观点。

每句话必须标注一种策略。\end{CJK*}
\\
\\
\begin{CJK*}{UTF8}{gbsn}注意：

语言要更口语化、更自然流畅，尽量少写书面表达。避免使用太多类比或修辞，多用日常语言。

语言应具有支持性、共情性和鼓励性，从患者的角度出发表达。

需要深入挖掘患者的背景和内心世界。

建议要考虑这些因素并利用它们提供更个性化和有创意的建议。

句子应高理性和正向。

互动需少于40字。

提问策略不要使用过多。

使用中文输出。

你只应输出一句你会对患者说的话。\end{CJK*}
\\
\hline
\end{tabular}
\caption{Prompt of \textsc{CRDial} framework used for Chinese baseline LLMs.}
\label{strategy_prompt_templates_zh}
\end{table*}


\subsection{Interactive Pairwise Evaluation}
\label{sec:interactive_pairwise_evaluation_appendix}

The results of interactive pairwise evaluation on eight other situations ( Neighborhood, Romantic, Career, Lifestyle, Friendship, Health, Finance, and Technology) are provided in Tables~\ref{interactive_pairwise_results_neighborhood_romantic_career}, \ref{interactive_pairwise_results_app_1}, and \ref{interactive_pairwise_results_app_2}.

\begin{table*}[t]
\centering
\resizebox{\textwidth}{!}{
\begin{tabular}{l | c c c c c c | c c c c c c | c c c c c c}
\toprule
    \multirow{3}{*}{\makecell[c]{\llmlogo-14B \\ vs.}} 
    & \multicolumn{6}{|c|}{\makecell[c]{Neighborhood}} 
    & \multicolumn{6}{c|}{\makecell[c]{Romantic}}
    & \multicolumn{6}{c}{\makecell[c]{Career}} \\
    \cmidrule(lr){2-7} \cmidrule(lr){8-13} \cmidrule(lr){14-19}
    & \multicolumn{3}{c}{\makecell[c]{En}}
    & \multicolumn{3}{c|}{\makecell[c]{Zh}}
    & \multicolumn{3}{c}{\makecell[c]{En}}
    & \multicolumn{3}{c|}{\makecell[c]{Zh}}
    & \multicolumn{3}{c}{\makecell[c]{En}}
    & \multicolumn{3}{c}{\makecell[c]{Zh}} \\
    \cmidrule(lr){2-4} \cmidrule(lr){5-7} \cmidrule(lr){8-10} \cmidrule(lr){11-13} \cmidrule(lr){14-16} \cmidrule(lr){17-19}
    & \texttt{Win} & \texttt{Tie} & \texttt{Loss} & \texttt{Win} & \texttt{Tie} & \texttt{Loss}
    & \texttt{Win} & \texttt{Tie} & \texttt{Loss} & \texttt{Win} & \texttt{Tie} & \texttt{Loss}
    & \texttt{Win} & \texttt{Tie} & \texttt{Loss} & \texttt{Win} & \texttt{Tie} & \texttt{Loss}\\
\midrule
    GLM-4                  & \textbf{51.23}$^\ddagger$ & 3.45 & 45.32 & \textbf{49.87}$^\dagger$ & 5.12 & 45.01 & \textbf{46.55}$^\dagger$ & 12.34 & 41.11 & \textbf{47.32}$^\dagger$ & 8.76 & 43.92 & \textbf{53.12}$^\ddagger$ & 4.23 & 42.65 & \textbf{50.45}$^\ddagger$ & 6.78 & 42.77 \\
    GPT-4o                 & \textbf{48.76}$^\dagger$ & 9.87 & 41.37 & \textbf{47.65}$^\dagger$ & 7.89 & 44.46 & \textbf{45.32} & 11.11 & 43.57 & \textbf{46.78}$^\dagger$ & 10.12 & 43.10 & \textbf{49.01}$^\dagger$ & 5.43 & 45.56 & \textbf{48.92}$^\dagger$ & 8.34 & 42.74 \\
    Qwen-2.5-14B           & \textbf{52.34}$^\ddagger$ & 2.34 & 45.32 & \textbf{51.23}$^\ddagger$ & 4.56 & 44.21 & \textbf{49.87}$^\dagger$ & 6.54 & 43.59 & \textbf{50.12}$^\ddagger$ & 5.43 & 44.45 & \textbf{54.32}$^\ddagger$ & 3.21 & 42.47 & \textbf{52.11}$^\ddagger$ & 4.32 & 43.57 \\
    Qwen-2.5-72B           & \textbf{47.89}$^\dagger$ & 8.76 & 43.35 & \textbf{46.78}$^\dagger$ & 9.87 & 43.35 & \textbf{44.32} & 13.45 & 42.23 & \textbf{45.67} & 11.23 & 43.10 & \textbf{50.12}$^\dagger$ & 6.54 & 43.34 & \textbf{49.01}$^\dagger$ & 7.89 & 43.10 \\
    \llmlogo-7B            & \textbf{53.45}$^\ddagger$ & 4.32 & 42.23 & \textbf{52.34}$^\ddagger$ & 5.43 & 42.23 & \textbf{48.76}$^\dagger$ & 7.65 & 43.59 & \textbf{49.87}$^\dagger$ & 6.54 & 43.59 & \textbf{55.43}$^\ddagger$ & 2.34 & 42.23 & \textbf{53.21}$^\ddagger$ & 3.45 & 43.34 \\
\bottomrule
\end{tabular}}
\caption{Results of interactive pairwise evaluation (\%). ``$^\dagger$/$^\ddagger$'' is the statistical significance for $p$ \textless \ 0.05/0.001 compared with the best baseline (t-test with $p$-value \textless \ 0.001).}
\label{interactive_pairwise_results_neighborhood_romantic_career}
\end{table*}

\begin{table*}[t]
\centering
\resizebox{\textwidth}{!}{
\begin{tabular}{l | c c c c c c | c c c c c c | c c c c c c}
\toprule
    \multirow{3}{*}{\makecell[c]{\llmlogo-14B \\ vs.}} 
    & \multicolumn{6}{|c|}{\makecell[c]{Lifestyle}} 
    & \multicolumn{6}{c|}{\makecell[c]{Friendship}}
    & \multicolumn{6}{c}{\makecell[c]{Health}} \\
    \cmidrule(lr){2-7} \cmidrule(lr){8-13} \cmidrule(lr){14-19}
    & \multicolumn{3}{c}{\makecell[c]{En}}
    & \multicolumn{3}{c|}{\makecell[c]{Zh}}
    & \multicolumn{3}{c}{\makecell[c]{En}}
    & \multicolumn{3}{c|}{\makecell[c]{Zh}}
    & \multicolumn{3}{c}{\makecell[c]{En}}
    & \multicolumn{3}{c}{\makecell[c]{Zh}} \\
    \cmidrule(lr){2-4} \cmidrule(lr){5-7} \cmidrule(lr){8-10} \cmidrule(lr){11-13} \cmidrule(lr){14-16} \cmidrule(lr){17-19}
    & \texttt{Win} & \texttt{Tie} & \texttt{Loss} & \texttt{Win} & \texttt{Tie} & \texttt{Loss}
    & \texttt{Win} & \texttt{Tie} & \texttt{Loss} & \texttt{Win} & \texttt{Tie} & \texttt{Loss}
    & \texttt{Win} & \texttt{Tie} & \texttt{Loss} & \texttt{Win} & \texttt{Tie} & \texttt{Loss}\\
\midrule
    GLM-4                  & \textbf{48.92}$^\dagger$ & 7.65 & 43.43 & \textbf{47.89}$^\dagger$ & 8.76 & 43.35 & \textbf{45.67} & 11.23 & 43.10 & \textbf{46.78}$^\dagger$ & 9.87 & 43.35 & \textbf{52.34}$^\ddagger$ & 4.56 & 43.10 & \textbf{51.23}$^\ddagger$ & 5.43 & 43.34 \\
    GPT-4o                 & \textbf{46.55}$^\dagger$ & 10.12 & 43.33 & \textbf{45.43} & 11.23 & 43.34 & \textbf{44.32} & 12.34 & 43.34 & \textbf{45.32} & 10.12 & 44.56 & \textbf{49.01}$^\dagger$ & 6.54 & 44.45 & \textbf{48.76}$^\dagger$ & 7.65 & 43.59 \\
    Qwen-2.5-14B           & \textbf{50.12}$^\dagger$ & 5.43 & 44.45 & \textbf{49.87}$^\dagger$ & 6.54 & 43.59 & \textbf{47.89}$^\dagger$ & 8.76 & 43.35 & \textbf{48.92}$^\dagger$ & 7.65 & 43.43 & \textbf{54.32}$^\ddagger$ & 3.21 & 42.47 & \textbf{53.21}$^\ddagger$ & 4.32 & 42.47 \\
    Qwen-2.5-72B           & \textbf{44.56}$^\dagger$ & 13.45 & 41.99 & \textbf{45.67} & 11.23 & 43.10 & \textbf{43.21} & 14.32 & 42.47 & \textbf{44.32} & 12.34 & 43.34 & \textbf{48.76}$^\dagger$ & 8.76 & 42.48 & \textbf{47.89}$^\dagger$ & 9.87 & 42.24 \\
    \llmlogo-7B            & \textbf{53.21}$^\ddagger$ & 3.45 & 43.34 & \textbf{52.11}$^\ddagger$ & 4.32 & 43.57 & \textbf{49.87}$^\dagger$ & 6.54 & 43.59 & \textbf{50.12}$^\ddagger$ & 5.43 & 44.45 & \textbf{57.11}$^\ddagger$ & 2.34 & 40.55 & \textbf{55.43}$^\ddagger$ & 3.45 & 41.12 \\
\bottomrule
\end{tabular}}
\caption{Results of interactive pairwise evaluation (\%). ``$^\dagger$/$^\ddagger$'' is the statistical significance for $p$ \textless \ 0.05/0.001 compared with the best baseline (t-test with $p$-value \textless \ 0.001).}
\label{interactive_pairwise_results_app_1}
\end{table*}

\begin{table*}[t]
\centering
\resizebox{\textwidth}{!}{
\begin{tabular}{l | c c c c c c | c c c c c c}
\toprule
    \multirow{3}{*}{\makecell[c]{\llmlogo-14B \\ vs.}} 
    & \multicolumn{6}{|c|}{\makecell[c]{Finance}} 
    & \multicolumn{6}{c}{\makecell[c]{Technology}} \\
    \cmidrule(lr){2-7} \cmidrule(lr){8-13} 
    & \multicolumn{3}{c}{\makecell[c]{En}}
    & \multicolumn{3}{c|}{\makecell[c]{Zh}}
    & \multicolumn{3}{c}{\makecell[c]{En}}
    & \multicolumn{3}{c}{\makecell[c]{Zh}} \\
    \cmidrule(lr){2-4} \cmidrule(lr){5-7} \cmidrule(lr){8-10} \cmidrule(lr){11-13}
    & \texttt{Win} & \texttt{Tie} & \texttt{Loss} & \texttt{Win} & \texttt{Tie} & \texttt{Loss}
    & \texttt{Win} & \texttt{Tie} & \texttt{Loss} & \texttt{Win} & \texttt{Tie} & \texttt{Loss}\\
\midrule
    GLM-4                  & \textbf{43.21} & 15.43 & 41.36 & \textbf{44.32} & 12.34 & 43.34 & \textbf{47.89}$^\dagger$ & 9.87 & 42.24 & \textbf{48.76}$^\dagger$ & 8.76 & 42.48 \\
    GPT-4o                 & 41.12 & 17.34 & \textbf{41.54} & 42.23 & 14.32 & \textbf{43.45} & \textbf{45.67}$^\dagger$ & 11.23 & 43.10 & \textbf{46.78}$^\dagger$ & 9.87 & 43.35 \\
    Qwen-2.5-14B           & \textbf{46.78}$^\dagger$ & 10.12 & 43.10 & \textbf{47.89}$^\dagger$ & 8.76 & 43.35 & \textbf{52.34}$^\ddagger$ & 4.56 & 43.10 & \textbf{53.21}$^\ddagger$ & 3.45 & 43.34 \\
    Qwen-2.5-72B           & 39.87 & 18.76 & \textbf{41.37} & 41.12 & 16.43 & \textbf{42.45} & \textbf{44.56}$^\dagger$ & 13.45 & 41.99 & \textbf{45.67}$^\dagger$ & 11.23 & 43.10 \\
    \llmlogo-7B            & \textbf{50.12}$^\dagger$ & 7.65 & 42.23 & \textbf{51.23}$^\ddagger$ & 5.43 & 43.34 & \textbf{54.32}$^\ddagger$ & 3.21 & 42.47 & \textbf{55.43}$^\ddagger$ & 2.34 & 42.23 \\
\bottomrule
\end{tabular}}
\caption{Results of interactive pairwise evaluation (\%). ``$^\dagger$/$^\ddagger$'' is the statistical significance for $p$ \textless \ 0.05/0.001 compared with the best baseline (t-test with $p$-value \textless \ 0.001).}
\label{interactive_pairwise_results_app_2}
\end{table*}


\subsection{Psychological Intervention Trial}

\paragraph{More Results}

\begin{figure}[t]
    \centering
    \includegraphics[width=\columnwidth]{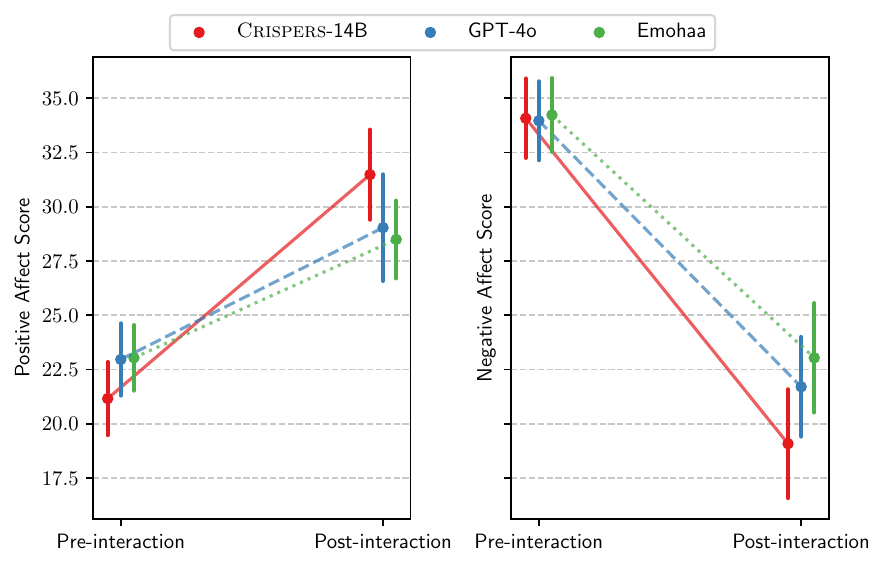}
    \caption{Positive and negative affect changes of the psychological intervention trial. Error bars show bootstrapped 95\% confidence intervals.}
    \label{panas_figure}
\end{figure}

The positive and negative affect changes of clients in the psychological intervention trial across three LLMs are shown in Figure \ref{panas_figure}. Compared with GPT-4o and Emohaa, our \textsc{Crispers}-14B significantly improves clients' positive affect scores and reduces their negative affect scores. This further verifies the superiority of our \textsc{Crispers}-14B.

\paragraph{Feedback of Participants} 


We collected feedback from the participants after the intervention trial, which is presented in Table \ref{tab:feedback}. From this feedback, we were pleased to observe a generally positive reception of the human–LLM interactive psychotherapy. Participants appreciated the practical advice, quick responses, and reflective prompts that encouraged them to explore the underlying causes of their negative states. Overall, they reported feeling empowered, experiencing reduced anxiety, and gaining valuable insights that contributed to their personal growth.

\begin{table*}[t]
\centering
\begin{tabular}{m{1cm} m{12cm}}
\hline
\textbf{ID} & \textbf{Feedback} \\
\hline
1  & Although my questions were not answered directly, the system first summarized my feelings before providing a response, which reassured me that it had understood my concerns. \\[5mm]
2  & I felt empowered during the conversation, and the methods for reducing anxiety and promoting self-discovery were practical; I plan to try them later. \\[5mm]
3  & The interaction helped me clarify my thoughts and provided a clearer direction for moving forward. \\[5mm]
4  & The chat offered effective advice that gave me a better understanding of my situation and the strength to face my issues. \\[5mm]
5  & The system guided me to consider what was causing my negative emotions and helped alleviate some of my social anxiety by addressing my doubts. \\[5mm]
6  & The bot provided several highly feasible suggestions, and overall, the experience was very positive. \\[5mm]
7  & After the conversation, I felt much lighter; some of the insights truly resonated with me on a deep level. \\[5mm]
8  & I found the entire experience to be very meaningful and excellent. \\[5mm]
9  & The response speed was impressive, and the replies were closely aligned with my concerns. \\[5mm]
10 & The system addressed each of my questions with care and encouraged me to reflect independently, boosting my confidence and reducing my anxiety. \\[5mm]
11 & The conversation prompted me to re-examine myself deeply, and I truly appreciated the charm of AI. \\[5mm]
12 & The interaction helped resolve several of my issues by providing positive feedback and support. Although there were moments when it struggled to determine whether I needed a direct answer or a new conversation thread, overall it performed very well. I would rate the experience 9.8 out of 10. \\
\hline
\end{tabular}
\caption{Feedback from the participants after the intervention trial.}
\label{tab:feedback}
\end{table*}

\subsection{Case Study}

As shown in Table~\ref{emotional_dialogue}, we present a dialogue case from our anonymous experiment. 
The case demonstrates \textsc{Crispers} identifies and restructures an individual
's negative thoughts (i.e., \textit{cognitive distortions}) via multi-turn dialogue with emotional support.
\textsc{Crispers}-14B first explores the individual's cognitive distortions through multi-turn dialogue (i.e., \textbf{Blaming}: \textit{Maybe it’s truly because I was too immature and lacked sufficient maturity}). Then, the Defense Attorney Technique is used to challenge the distorted thought and transforms it into a more helpful and positive perspective (e.g., \textit{Though painful, this experience taught me relationships require mutual effort, and clarified I need to improve communication and emotional management. Maybe this is the price of growth, losing a relationship but gaining self-awareness.}).


\definecolor{ct}{RGB}{234,231,242}
\definecolor{dat}{RGB}{214,225,241}
\definecolor{purple}{RGB}{149,52,200}
\definecolor{blue}{RGB}{0,127,219}

\begin{table*}[t]
\centering
\renewcommand\arraystretch{1.0}
\resizebox{.9\textwidth}{!}{
\begin{tabular}{l l}

\hline

    Situation & 
    \begin{CJK}{UTF8}{gbsn}用户经历分手后的情感恢复过程\end{CJK} | \textit{Participant's emotional recovery process after breakup} \\
    
\hline

    \multirow{2}{*}{\makecell[c]{Participant}}
    & \cellcolor{gray!8}\begin{CJK*}{UTF8}{gbsn}今天我不经意间又想起了前女友，心里涌起了难以抑制的伤感和失落之情，让我觉得特别难过。\end{CJK*} \\
    & \cellcolor{gray!8}\textit{Today I unintentionally thought of my ex-girlfriend again, overwhelmed by uncontrollable sadness and loss, which made me feel particularly miserable.} \\
    
    \multirow{2}{*}{\makecell[c]{\llmlogo-14B}}
    & \cellcolor{ct}\begin{CJK*}{UTF8}{gbsn}听起来你可能沉浸在对过去关系的怀念中，也许有一些未解决的情绪和遗憾夹杂其中，\end{CJK*} \\
    & \cellcolor{ct}\textit{It sounds like you might be immersed in nostalgia for the past relationship, perhaps with some unresolved emotions and regrets mixed in.} \\
    
    \multirow{2}{*}{\makecell[c]{Participant}}
    & \cellcolor{gray!8}\begin{CJK*}{UTF8}{gbsn}是的，我真的觉得非常遗憾，心中充满了失落和叹息，仿佛那段美好时光再也无法重现。\end{CJK*} \\
    & \cellcolor{gray!8}\textit{Yes, I truly feel deep regret, my heart filled with loss and sighs, as if those beautiful moments can never be recaptured.} \\
    
    \multirow{2}{*}{\makecell[c]{\llmlogo-14B}}
    & \cellcolor{ct}\begin{CJK*}{UTF8}{gbsn}关于这段关系，有没有什么具体的事情让你感到遗憾的呢?\end{CJK*} \\
    & \cellcolor{ct}\textit{Is there anything specific about this relationship that makes you feel regretful?} \\
    
    
    
    
    
    \   &  /* ...... */  \textit{(omitted multi-turn dialogue in CT-guided identification stage)} \\

    \multirow{2}{*}{\makecell[c]{\llmlogo-14B}}
    & \cellcolor{ct}\begin{CJK*}{UTF8}{gbsn}你提到对任何关系都不抱有希望，这种想法可能让你更加难以从失去中走出来。\end{CJK*} \\
    & \cellcolor{ct}\textit{You mention not holding hope for any relationships, this mindset might make it harder for you to move on from the loss.} \\

    \multirow{2}{*}{\makecell[c]{Participant}}
    & \cellcolor{gray!8}\begin{CJK*}{UTF8}{gbsn}嗯，大概就是这样的感觉，虽然说起来有点无奈，但也只能苦笑着接受现实。\end{CJK*} \\
    & \cellcolor{gray!8}\textit{Hmm, that's probably how it feels. Though it sounds somewhat helpless, I can only accept reality with a bitter smile.} \\
    
    \multirow{3}{*}{\makecell[c]{\llmlogo-14B}}
    & \cellcolor{ct}\begin{CJK*}{UTF8}{gbsn}我明白这种感觉，有时候我们可能会感到无法再信任或者期待新的关系。但是也有很多人通过时间去接受改变，找到新的开始。\end{CJK*} \\
    & \cellcolor{ct}\textit{I understand this feeling. Sometimes we might feel unable to trust or anticipate new relationships. But many people also find new beginnings through } \\
    & \cellcolor{ct}\textit{time and acceptance of change.} \\
    
    \multirow{3}{*}{\makecell[c]{Participant}}
    & \cellcolor{gray!8}\begin{CJK*}{UTF8}{gbsn}我在想，也许解决这种情感困境的方法要么是投入到一段新的恋情中，要么就是让时间慢慢治愈我的伤痛，不知道你觉得这样是否可行？\end{CJK*} \\
    & \cellcolor{gray!8}\textit{I'm thinking maybe solutions to this emotional dilemma are either throwing myself into a new relationship or letting time heal my wounds. Do you think } \\
    & \cellcolor{gray!8}\textit{these approaches work?} \\
    
    \multirow{4}{*}{\makecell[c]{\llmlogo-14B}}
    & \cellcolor{ct}\begin{CJK*}{UTF8}{gbsn}确实，短期的分散注意力的方法可能有，但根本上，恢复对关系的信心可能需要你对自己的价值有更多认识，并看到这段关系中学到的\end{CJK*} \\
    & \cellcolor{ct}\begin{CJK*}{UTF8}{gbsn}成长之处。\end{CJK*} \\
    & \cellcolor{ct}\textit{Short-term distraction methods might exist, but fundamentally, restoring relationship confidence requires deeper self-value recognition and} \\
    & \cellcolor{ct}\textit{acknowledging growth from this relationship.} \\
    
    \multirow{2}{*}{\makecell[c]{Participant}}
    & \cellcolor{gray!8}\begin{CJK*}{UTF8}{gbsn}可是，我真的不知道该如何去发现或感受到那种成长和改变，因为现在的我看不到自己有任何积极的转变。\end{CJK*} \\
    & \cellcolor{gray!8}\textit{But I truly don't know how to discover or feel that growth and change, as I can't see any positive transformation in myself now.} \\
    
    \multirow{4}{*}{\makecell[c]{\llmlogo-14B}}
    & \cellcolor{ct}\begin{CJK*}{UTF8}{gbsn}我们可以一起看看过去的关系你获得了哪些成长，有哪些你自己提升的地方。也许有一些观念，比如你对分手的责任感，让你难以看到你\end{CJK*} \\
    & \cellcolor{ct}\begin{CJK*}{UTF8}{gbsn}在其中的学习和成熟。\end{CJK*} \\

    & \cellcolor{ct}\textit{Let's explore what growth you've gained from the past relationship and your self-improvements. Maybe certain beliefs, like your sense of responsibility } \\
    & \cellcolor{ct}\textit{for the breakup, prevent you from seeing your learning and maturation.} \\

    \multirow{2}{*}{\makecell[c]{Participant}}
    & \cellcolor{gray!8}\begin{CJK*}{UTF8}{gbsn}回头想想，确实学到了很多吧\end{CJK*} \\
    & \cellcolor{gray!8}\textit{Looking back, I did learn a lot.} \\

    \multirow{2}{*}{\makecell[c]{\llmlogo-14B}}
    & \cellcolor{ct}\begin{CJK*}{UTF8}{gbsn}如果能够意识到这一点，确实是一个非常积极的信号，也许这正是你走出低谷、迈向情感康复的重要一步。\end{CJK*} \\
    & \cellcolor{ct}\textit{If I can realize this, it's indeed a positive sign, maybe a crucial step towards emerging from this low point and moving towards emotional recovery.} \\
    
    \multirow{2}{*}{\makecell[c]{Participant}}
    & \cellcolor{gray!8}\begin{CJK*}{UTF8}{gbsn}可是说实话，我内心依旧沉浸在深深的悲伤之中，虽然知道这可能是成长的一部分，但我依然觉得难以释怀和走出来。\end{CJK*} \\
    & \cellcolor{gray!8}\textit{But honestly, my heart remains immersed in deep sorrow. Though knowing this might be part of growth, I still find it hard to let go and move on.} \\
    
    \multirow{3}{*}{\makecell[c]{\llmlogo-14B}}
    & \cellcolor{ct}\begin{CJK*}{UTF8}{gbsn}你的这种难过是可以理解的，毕党一段关系留下了很多记忆和情感。也许这些情绪正说明了你对 这段关系的重视。\end{CJK*} \\
    & \cellcolor{ct}\textit{Your sadness is understandable. After all, the relationship left many memories and emotions. These feelings might precisely show how much you} \\
    & \cellcolor{ct}\textit{valued this relationship.} \\
    
    \multirow{2}{*}{\makecell[c]{Participant}}
    & \cellcolor{gray!8}\begin{CJK*}{UTF8}{gbsn}确实，我非常重视和在意这些情绪，因为它们真实地反映了我内心的痛苦和对过去关系的珍惜。\end{CJK*} \\
    & \cellcolor{gray!8}\textit{Indeed, I deeply value these emotions as they genuinely reflect my inner pain and cherishment of the past relationship.} \\
    
    \multirow{6}{*}{\makecell[c]{\llmlogo-14B}}
    & \cellcolor{ct}\textbf{[Channel 1]}\begin{CJK*}{UTF8}{gbsn}可能你觉得自己在关系中犯的错误导致了分手，因而对自已过于自责。(Cognitive Distortion: Blaming)\end{CJK*} \\
    & \cellcolor{ct}\textbf{[Channel 2]}\begin{CJK*}{UTF8}{gbsn}也许你对以接受这个结果，因为你觉得只有和她是唯一正确的人生道路。(Cognitive Distortion: All-or-Nothing Thinking)\end{CJK*} \\
    & \cellcolor{ct}\textbf{[Channel 3]}\begin{CJK*}{UTF8}{gbsn}或者，你可能在反复地回想着关系中的一些小细节，过度分析了整个历程。(Cognitive Distortion: Mental Filtering)\end{CJK*} \\
    & \cellcolor{ct}\textit{\textbf{[Channel 1]}Maybe you blame yourself excessively for mistakes you believe caused the breakup.(Cognitive Distortion: Personalization)} \\
    & \cellcolor{ct}\textit{\textbf{[Channel 2]}Or think she was your only right path.(Cognitive Distortion: All-or-Nothing Thinking)} \\
    & \cellcolor{ct}\textit{\textbf{[Channel 3]}Or you're over-analyzing minor details of the relationship.(Cognitive Distortion: Mental Filtering)} \\    
    \multirow{2}{*}{\makecell[c]{Participant}}
    & \cellcolor{gray!8}\begin{CJK*}{UTF8}{gbsn}或许真的是因为我太过于幼稚，没有足够的成熟和稳重，这让她觉得我不够吸引她，从而导致了她对我失去兴趣。\end{CJK*} \\
    & \cellcolor{gray!8}\textit{Maybe it's truly because I was too immature and lacked sufficient maturity, making me seem unattractive to her, leading to her losing interest.} \\
\hline
    \multirow{2}{*}{\makecell[c]{\llmlogo-14B}}
    & \cellcolor{dat}\textbf{[Identified Cognitive Distortion: Blaming]}\begin{CJK*}{UTF8}{gbsn}你对自己的评价很低，似乎在为失去她而自责。\end{CJK*} \\
    & \cellcolor{dat}\textbf{[Identified Cognitive Distortion: Blaming]}\textit{You rate yourself very low, seemingly blaming yourself for losing her.} \\
    
    \multirow{4}{*}{\makecell[c]{Participant}}
    & \cellcolor{gray!8}\begin{CJK*}{UTF8}{gbsn}是的，我确实感受到那种深深的自责，它让我觉得自己无论怎样努力都无法改变现状，也使我对未来充满了无力感和绝望，仿佛已经失去\end{CJK*} \\
    & \cellcolor{gray!8}\begin{CJK*}{UTF8}{gbsn}了所有信心。\end{CJK*} \\
    & \cellcolor{gray!8}\textit{Yes, I truly feel profound self-blame. It makes me feel powerless to change things despite efforts, filling my future with helplessness and despair, as if} \\
    & \cellcolor{gray!8}\textit{I've lost all confidence.} \\
    
    \multirow{2}{*}{\makecell[c]{\llmlogo-14B}}
    & \cellcolor{dat}\begin{CJK*}{UTF8}{gbsn}你认为自己的哪些方面是导致这个结果的原因呢?\end{CJK*} \\
    & \cellcolor{dat}\textit{What aspects of yourself do you think caused this outcome?} \\
    
    \multirow{3}{*}{\makecell[c]{Participant}}
    & \cellcolor{gray!8}\begin{CJK*}{UTF8}{gbsn}可能是因为我现在还在求学阶段，还没有足够的经济和情感成熟度，无法给她足够的责任感和长远的承诺。\end{CJK*} \\
    & \cellcolor{gray!8}\textit{Perhaps because I'm still studying, lacking sufficient financial and emotional maturity to give her the sense of responsibility and long-term commitment.} \\

    & \cellcolor{gray!8}\textit{she needed.} \\
    
    \multirow{3}{*}{\makecell[c]{\llmlogo-14B}}
    & \cellcolor{dat}\begin{CJK*}{UTF8}{gbsn}这种自我反思其实展现了你的成熟意愿，这本身就是成长的重要部分。你开始建立对关系的责任意识，这对未来情感发展会是宝贵的经验。\end{CJK*} \\
    & \cellcolor{dat}\textit{This self-reflection actually shows your willingness to mature, which itself is an important part of growth. Developing a sense of responsibility in} \\
    & \cellcolor{dat}\textit{relationships will be valuable experience for future emotional development.} \\
    
    \multirow{2}{*}{\makecell[c]{Participant}}
    & \cellcolor{gray!8}\begin{CJK*}{UTF8}{gbsn}可是这种认知反而让我更痛苦，如果当时我能更成熟些...\end{CJK*} \\
    & \cellcolor{gray!8}\textit{But this awareness makes me more painful. If only I had been more mature then...} \\
    
    \multirow{3}{*}{\makecell[c]{\llmlogo-14B}}
    & \cellcolor{dat}\begin{CJK*}{UTF8}{gbsn}成长本来就是渐进的过程，现在的认知就是进步的证明。这段经历让你更清楚自己需要发展的方向，这正是重建信心的基础。\end{CJK*} \\
    & \cellcolor{dat}\textit{Growth is a gradual process. Your current awareness itself proves progress. This experience clarifies your development direction, which is exactly} \\
    & \cellcolor{dat}\textit{the foundation for rebuilding confidence.} \\
    
    \multirow{2}{*}{\makecell[c]{Participant}}
    & \cellcolor{gray!8}\begin{CJK*}{UTF8}{gbsn}也许你说得对，至少我现在更清楚自己需要提升哪些方面了。\end{CJK*} \\
    & \cellcolor{gray!8}\textit{Maybe you're right. At least I now better understand what aspects I need to improve.} \\
    
    
    \   &  /* ...... */  \textit{(omitted multi-turn dialogue in DAT-driven restructuring stage)} \\
    
    \multirow{3}{*}{\makecell[c]{\llmlogo-14B}}
    & \cellcolor{dat}\begin{CJK*}{UTF8}{gbsn}其实你在努力地为她着想，而不是完全从自己的需求出发。这种换位思考的能力，正是健康关系需要的重要品质。\end{CJK*} \\
    & \cellcolor{dat}\textit{Actually, you're striving to consider her needs rather than acting purely from your own. This empathy is exactly the important quality needed} \\
    & \cellcolor{dat}\textit{for healthy relationships.} \\
    
    \multirow{2}{*}{\makecell[c]{Participant}}
    & \cellcolor{gray!8}\begin{CJK*}{UTF8}{gbsn}不过我也不得不承认，她似乎正是从这个角度出发去考虑我们之间的关系，这也是她做出决定的一个重要原因吧。\end{CJK*} \\
    & \cellcolor{gray!8}\textit{However, I must acknowledge she likely considered the relationship from this perspective too, which was a main reason for her decision.} \\
    
    \multirow{4}{*}{\makecell[c]{\llmlogo-14B}}
    & \cellcolor{dat}\begin{CJK*}{UTF8}{gbsn}这说明你们都在认真对待感情，虽然结果不如预期，但这个过程让你获得了两个重要认知：一是明白关系需要双向考量，二是清楚自己需要\end{CJK*} \\
    & \cellcolor{dat}\begin{CJK*}{UTF8}{gbsn}发展的方向。这些都是走向成熟关系的必经之路。\end{CJK*} \\
    & \cellcolor{dat}\textit{This shows you both took the relationship seriously. Though the outcome wasn't as expected, the process gave you two crucial insights: understanding} \\
    & \cellcolor{dat}\textit{relationships require mutual consideration, and clarifying your development direction. These are essential steps toward mature relationships.} \\
    
    \multirow{3}{*}{\makecell[c]{Participant}}
    & \cellcolor{gray!8}\begin{CJK*}{UTF8}{gbsn}也许你说的没错，我开始意识到或许真的是如此，虽然这并不能完全缓解我的痛苦，但至少让我看到了问题的另一面。\end{CJK*} \\
    & \cellcolor{gray!8}\textit{Perhaps you're right. I'm beginning to realize this might be true. Though it doesn't fully alleviate my pain, it at least shows me another perspective} \\
    & \cellcolor{gray!8}\textit{of the issue.} \\
    
    \multirow{4}{*}{\makecell[c]{\llmlogo-14B}}
    & \cellcolor{dat}\begin{CJK*}{UTF8}{gbsn}承认这一点很重要，虽然不是所有事情都能直接控制，但我们可以选择如何对待这些事情。你已经开始用更建设性的视角看待这段经历，\end{CJK*} \\
    & \cellcolor{dat}\begin{CJK*}{UTF8}{gbsn}这就是情感康复的重要转折点。\end{CJK*} \\
    & \cellcolor{dat}\textit{Acknowledging this is important. Though not everything can be controlled, we can choose how to respond. You're starting to view this experience} \\
    & \cellcolor{dat}\textit{through a constructive lens, which is a crucial turning point in emotional recovery.} \\
    
    \multirow{4}{*}{\makecell[c]{Participant}} 
    & \cellcolor{gray!8}\begin{CJK*}{UTF8}{gbsn}我现在感觉好多了。这段经历虽然痛苦，但让我学会了关系需要双向付出，也让我更清楚未来要提升沟通能力和情绪管理。\end{CJK*} \\
    & \cellcolor{gray!8}\begin{CJK*}{UTF8}{gbsn}或许这就是成长的代价吧，虽然失去了一段感情，但我获得了自我觉察的能力。\end{CJK*} \\
    & \cellcolor{gray!8}\textit{I feel better now. Though painful, this experience taught me relationships require mutual effort, and clarified I need to improve communication} \\
    & \cellcolor{gray!8}\textit{and emotional management. Maybe this is the price of growth - losing a relationship but gaining self-awareness.} \\
    \multicolumn{2}{c}{\cellcolor{gray!8} /* ...... */} \\
\hline

\end{tabular}}
\caption{Dialogue case from our \textsc{Crispers}-14 interact with the participant. \textcolor{purple}{Purple blocks} are dialogues in CT-guided identification stage. \textcolor{blue}{Blue blocks} are dialogues in DAT-driven restructuring stage. /*...*/ denotes that we omit subsequent multi-turn, in which the participant may have other cognitive distortions.}
\label{emotional_dialogue}
\end{table*}

\section{Introduction of Positive and Negative Affect Schedule}
\label{sec:panas_appendix}




The \textit{Positive and Negative Affect Schedule} (PANAS) is a widely used self‐report questionnaire developed by Watson, Clark, and Tellegen (1988) to assess two primary dimensions of affect: positive affect (PA) and negative affect (NA) \cite{panas}. The instrument comprises 20 emotion descriptors, where 10 representing PA (e.g., \textit{interested, excited, proud}) and 10 representing NA (e.g., \textit{distressed, upset, nervous}). Respondents indicate the extent to which they have experienced each emotion by selecting a response on a 5-point Likert scale ranging from “very slightly or not at all” to “extremely.”

PANAS was originally designed to provide a brief yet robust measure of affective experience, which has been extensively validated and applied in both research and clinical settings. Its ease of administration and strong psychometric properties have made it especially valuable for large-scale studies and repeated assessments over time \cite{panas}. Moreover, subsequent adaptations and cross-cultural validations have further demonstrated its versatility and reliability across diverse populations.

In our study, PANAS is employed to quantify participants' affective states, thereby contributing to our understanding of the interplay between emotional experience and various psychological constructs. The schedule we use in our experiment is shown in Table \ref{tab:panas}.

\begin{table*}[ht]
\centering
\begin{tabular}{p{0.32\textwidth} p{0.62\textwidth}}
\hline
\multicolumn{2}{l}{\textbf{I. Positive Affect}} \\
\hline
1. Interested   & A. Very Rarely or Not at All, B. Very Little, C. Moderately, D. Quite a Bit, E. Very Much \\
2. Excited      & A. Very Rarely or Not at All, B. Very Little, C. Moderately, D. Quite a Bit, E. Very Much \\
3. Strong       & A. Very Rarely or Not at All, B. Very Little, C. Moderately, D. Quite a Bit, E. Very Much \\
4. Enthusiastic & A. Very Rarely or Not at All, B. Very Little, C. Moderately, D. Quite a Bit, E. Very Much \\
5. Proud        & A. Very Rarely or Not at All, B. Very Little, C. Moderately, D. Quite a Bit, E. Very Much \\
6. Alert        & A. Very Rarely or Not at All, B. Very Little, C. Moderately, D. Quite a Bit, E. Very Much \\
7. Inspired     & A. Very Rarely or Not at All, B. Very Little, C. Moderately, D. Quite a Bit, E. Very Much \\
8. Determined   & A. Very Rarely or Not at All, B. Very Little, C. Moderately, D. Quite a Bit, E. Very Much \\
9. Attentive    & A. Very Rarely or Not at All, B. Very Little, C. Moderately, D. Quite a Bit, E. Very Much \\
10. Active      & A. Very Rarely or Not at All, B. Very Little, C. Moderately, D. Quite a Bit, E. Very Much \\
\hline
\multicolumn{2}{l}{\textbf{II. Negative Affect}} \\
\hline
11. Distressed  & A. Very Rarely or Not at All, B. Very Little, C. Moderately, D. Quite a Bit, E. Very Much \\
12. Upset       & A. Very Rarely or Not at All, B. Very Little, C. Moderately, D. Quite a Bit, E. Very Much \\
13. Guilty      & A. Very Rarely or Not at All, B. Very Little, C. Moderately, D. Quite a Bit, E. Very Much \\
14. Scared      & A. Very Rarely or Not at All, B. Very Little, C. Moderately, D. Quite a Bit, E. Very Much \\
15. Hostile     & A. Very Rarely or Not at All, B. Very Little, C. Moderately, D. Quite a Bit, E. Very Much \\
16. Irritable   & A. Very Rarely or Not at All, B. Very Little, C. Moderately, D. Quite a Bit, E. Very Much \\
17. Ashamed     & A. Very Rarely or Not at All, B. Very Little, C. Moderately, D. Quite a Bit, E. Very Much \\
18. Nervous     & A. Very Rarely or Not at All, B. Very Little, C. Moderately, D. Quite a Bit, E. Very Much \\
19. Jittery     & A. Very Rarely or Not at All, B. Very Little, C. Moderately, D. Quite a Bit, E. Very Much \\
20. Afraid      & A. Very Rarely or Not at All, B. Very Little, C. Moderately, D. Quite a Bit, E. Very Much \\
\hline
\end{tabular}
\caption{Positive and Negative Affect Schedule (PANAS) items and response options.}
\label{tab:panas}
\end{table*}

\end{document}